\documentclass[final]{cvpr}

\usepackage{times}
\usepackage{epsfig}
\usepackage{graphicx}
\usepackage{amsmath}
\usepackage{amssymb}
\usepackage{amsfonts}
\usepackage{amsthm}
\usepackage{bm}
\usepackage{fixmath}
\usepackage{booktabs}
\usepackage{xspace}
\usepackage{tabularx}
\usepackage{cases}
\usepackage{subfigure}
\usepackage{indentfirst}
\usepackage{multirow}
\usepackage{boldline}
\usepackage{enumitem}
\usepackage{mathrsfs}
\usepackage{algorithmic}
\usepackage[table]{xcolor}
\usepackage{pifont}

\usepackage[pagebackref=true,breaklinks=true,colorlinks,bookmarks=false]{hyperref}
\usepackage{url}

\def\cvprPaperID{5629} 
\def\confYear{CVPR 2021}

\renewcommand{\vec}[1]{\boldsymbol{#1}}

\newcommand{\ournet}{MoViNet\xspace} 
\newcommand{\ournets}{\ournet{}s\xspace}

\newcommand{\eat}[1]{}

\newcommand{\appendixsearchspace}{\ref{appendix:search-space}}
\newcommand{\appendixarchitectures}{\ref{appendix:architectures}}
\newcommand{\appendixmoreresults}{\ref{appendix:more-results}}

\begin{document}

\title{\ournets: Mobile Video Networks for Efficient Video Recognition}

\author{
    Dan Kondratyuk\thanks{Work done as a part of the Google AI Residency.},
    Liangzhe Yuan,
    Yandong Li,
    Li Zhang,
    Mingxing Tan,
    Matthew Brown,
    Boqing Gong\\
    Google Research\\
    {\tt\small \{dankondratyuk,lzyuan,yandongli,zhl,tanmingxing,mtbr,bgong\}@google.com}
}

\maketitle

\begin{abstract}
    We present Mobile Video Networks (\ournets), a family of computation and memory efficient video networks that can operate on streaming video for online inference.
    3D convolutional neural networks (CNNs) are accurate at video recognition but require large computation and memory budgets and do not support online inference, making them difficult to work on mobile devices.
    We propose a three-step approach to improve computational efficiency while substantially reducing the peak memory usage of 3D CNNs.
    First, we design a video network search space and employ neural architecture search to generate efficient and diverse 3D CNN architectures.
    Second, we introduce the Stream Buffer technique that decouples memory from video clip duration, allowing 3D CNNs to embed arbitrary-length streaming video sequences for both training and inference with a small constant memory footprint.
    Third, we propose a simple ensembling technique to improve accuracy further without sacrificing efficiency.
    These three progressive techniques allow \ournets to achieve state-of-the-art accuracy and efficiency on the Kinetics, Moments in Time, and Charades video action recognition datasets.
    For instance, \ournet-A5-Stream achieves the same accuracy as X3D-XL on Kinetics 600 while requiring 80\% fewer FLOPs and 65\% less memory.
    Code will be made available at \url{https://github.com/tensorflow/models/tree/master/official/vision}.
\end{abstract}

\vspace{-12pt}
\section{Introduction} \label{sec:intro}
\begin{figure}[t]
    \begin{center}
    \includegraphics[width=1.0\linewidth]{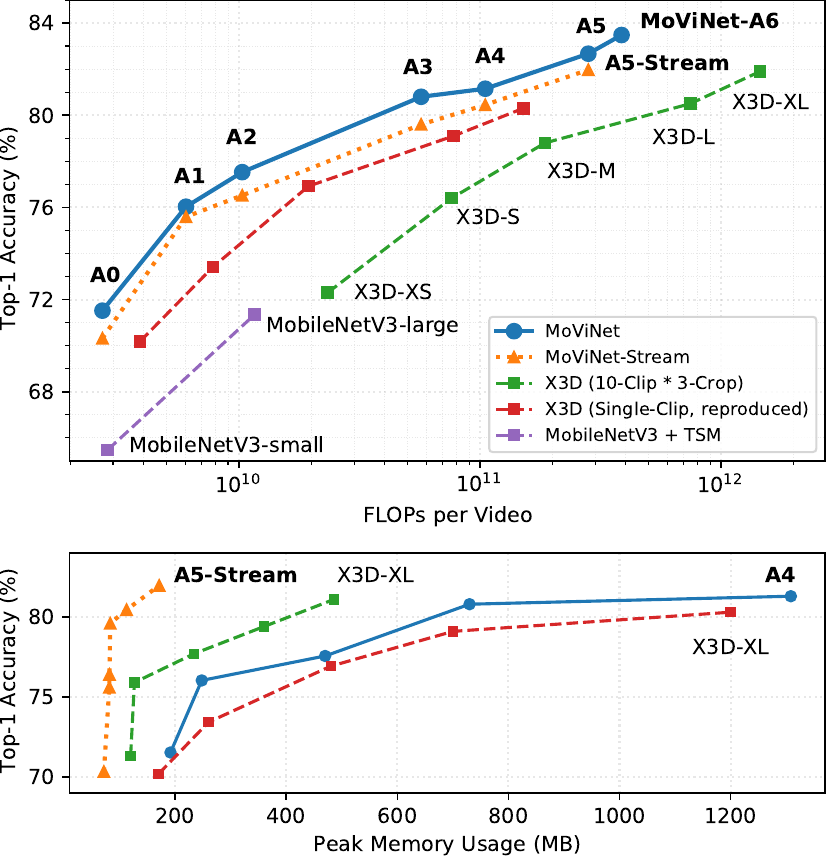}
    \end{center}
        \vspace{-5pt}
        \caption{
            {\bf Accuracy vs. FLOPs and Memory on Kinetics 600}.
            \ournets are more accurate than 2D networks and more efficient than 3D networks.
            Top (log scale): \ournet-A2 achieves {\bf 6\% higher} accuracy than MobileNetV3~\cite{howard2019searching} at the same FLOPs while \ournet-A6 achieves state-of-the-art 83.5\% accuracy being {\bf 5.1x faster} than X3D-XL~\cite{feichtenhofer2020x3d}.
            Bottom: Streaming \ournets require {\bf 10x less memory} at the cost of 1\% accuracy.
            Note that we only train on the ~93\% of Kinetics 600 examples that are available at the time of writing.
            Best viewed in color.
        }
        \vspace{-12pt}
    \label{fig:k600-comparison}
\end{figure}

Efficient video recognition models are opening up new opportunities for mobile camera, IoT, and self-driving applications where efficient and accurate on-device processing is paramount.
Despite recent advances in deep video modeling, it remains difficult to find models that run on mobile devices and achieve high video recognition accuracy.
On the one hand, 3D convolutional neural networks (CNNs)~\cite{tran2018closer, wang2018non, feichtenhofer2019slowfast, feichtenhofer2020x3d, ryoo2019assemblenet} offer state-of-the-art accuracy, but consume copious amounts of memory and computation.
On the other hand, 2D CNNs~\cite{lin2019tsm, zhu2020faster} require far fewer resources suitable for mobile and can run online using frame-by-frame prediction, but fall short in accuracy.

Many operations that make 3D video networks accurate (e.g., temporal convolution, non-local blocks~\cite{wang2018non}, etc.) require all input frames to be processed at once, limiting the opportunity for accurate models to be deployed on mobile devices.
The recently proposed X3D networks~\cite{feichtenhofer2020x3d} provide a significant effort to increase the efficiency of 3D CNNs.
However, they require large memory resources on large temporal windows which incur high costs, or small temporal windows which reduce accuracy.
Other works aim to improve 2D CNNs' accuracy using temporal aggregation~\cite{lin2019tsm, fan2019more, wu2020dynamic, liu2020tam, fan2020rubiksnet}, however their limited inter-frame interactions reduce these models' abilities to adequately model long-range temporal dependencies like 3D CNNs.

This paper introduces {\it three progressive steps} to design efficient video models which we use to produce Mobile Video Networks ({\bf \ournets}), a family of memory and computation efficient 3D CNNs.

\begin{enumerate}
    \item We first define a {\bf \ournet search space} to allow Neural Architecture Search (NAS) to efficiently trade-off spatiotemporal feature representations.
    \item We then introduce {\bf Stream Buffers} for \ournets, which process videos in small consecutive subclips, requiring constant memory without sacrificing long temporal dependencies, and which enable online inference.
    \item Finally, we create {\bf Temporal Ensembles} of streaming \ournets, regaining the slightly lost accuracy from the stream buffers.
\end{enumerate}

First, we design the \ournet search space to explore how to mix spatial, temporal, and spatiotemporal operations such that NAS can find optimal feature combinations to trade-off efficiency and accuracy.
Figure~\ref{fig:k600-comparison} visualizes the efficiency of the generated \ournets.
\ournet-A0 achieves similar accuracy to MobileNetV3-large+TSM~\cite{howard2019searching, lin2019tsm} on Kinetics 600~\cite{kay2017kinetics} with 75\% fewer FLOPs.
\ournet-A6 achieves state-of-the-art 83.5\% accuracy, 1.6\% higher than X3D-XL~\cite{feichtenhofer2020x3d}, requiring 60\% fewer FLOPs.

Second, we create streaming \ournets by introducing the stream buffer to reduce memory usage from linear to constant in the number of input frames for both training and inference, allowing \ournets to run with substantially fewer memory bottlenecks.
E.g., the stream buffer reduces \ournet-A5's memory usage by 90\%.
In contrast to traditional multi-clip evaluation (test-time data augmentation) approaches~\cite{simonyan2014two, wang2015towards} which also reduce memory, a stream buffer carries over temporal dependencies between consecutive non-overlapping subclips by caching feature maps at subclip boundaries.
The stream buffer allows for a larger class of operations to enhance online temporal modeling than the recently proposed temporal shift~\cite{lin2019tsm}.
We equip the stream buffer with temporally unidirectional causal operations like causal convolution~\cite{oord2016wavenet}, cumulative pooling, and causal squeeze-and-excitation~\cite{hu2018squeeze} with positional encoding to force temporal receptive fields to look only into past frames, enabling \ournets to operate incrementally on streaming video for online inference.
However, the causal operations come at a small cost, reducing accuracy on Kinetics 600 by 1\% on average.

Third, we temporally ensemble \ournets, showing that they are more accurate than single large networks while achieving the same efficiency.
We train two streaming \ournets independently with the same total FLOPs as a single model and average their logits.
This simple technique gains back the loss in accuracy when using stream buffers.

Taken together, these three techniques create \ournets that are high in accuracy, low in memory usage, efficient in computation, and support online inference.
We search for \ournets using the Kinetics 600 dataset~\cite{carreira2018short} and test them extensively on Kinetics 400~\cite{kay2017kinetics}, Kinetics 700~\cite{carreira2019short}, Moments in Time~\cite{monfort2019moments}, Charades~\cite{sigurdsson2016hollywood}, and Something-Something~V2~\cite{goyal2017something}.

\section{Related Work} \label{sec:related-work}
\paragraph{Efficient Video Modeling.}

Deep neural networks have made remarkable progress for video understanding~\cite{ji20123d,simonyan2014two,tran2015learning, wang2016temporal,carreira2017quo, wang2018non,qiu2019learning,feichtenhofer2020x3d,feichtenhofer2019slowfast}.
They extend 2D image models with a temporal dimension, most notably incorporating 3D convolution~\cite{ji20123d,taylor2010convolutional,tran2015learning,xie2017rethinking,hara2018can,qiu2017learning,jiang2019stm,ryoo2019assemblenet}.

Improving the efficiency of video models has gained increased attention~\cite{feichtenhofer2019slowfast,tran2019video,feichtenhofer2017spatiotemporal,feichtenhofer2020x3d,lin2019tsm,fan2019more,bhardwaj2019efficient,chen2018big,li2020smallbignet,piergiovanni2020tiny}.
Some works explore the use of 2D networks for video recognition by processing videos in smaller segments followed by late fusion~\cite{karpathy2014large,donahue2015long,yue2015beyond,wang2016temporal,feichtenhofer2017spatiotemporal,sun2017lattice,li2018recurrent,li2018videolstm,wang2018non,zhou2018temporal,zhu2020faster}. 
The Temporal Shift Module~\cite{lin2019tsm} uses early fusion to shift a portion of channels along the temporal axis, boosting accuracy while supporting online inference.


\vspace{-12pt}
\paragraph{Causal Modeling.}
WaveNet~\cite{oord2016wavenet} introduces causal convolution, where the receptive field of a stack of 1D convolutions only extends to features up to the current time step.
We take inspiration from other works using causal convolutions~\cite{carreira2018massively,chang2018temporal,dai2019transformer,cheng2019sparse,daiya2020stock} to design stream buffers for online video model inference, allowing frame-by-frame predictions with 3D kernels.

\vspace{-12pt}
\paragraph{Multi-Objective NAS.}
The use of NAS~\cite{zoph2016neural,liu2018progressive,pham2018efficient,tan2019mnasnet,cai2018proxylessnas,kandasamy2018neural} with multi-objective architecture search has also grown in interest, producing more efficient models in the process for image recognition~\cite{tan2019mnasnet, cai2018proxylessnas, bender2020can} and video recognition~\cite{piergiovanni2020tiny, ryoo2019assemblenet}.
We make use of TuNAS~\cite{bender2020can}, a one-shot NAS framework which uses aggressive weight sharing that is well-suited for computation intensive video models.

\vspace{-12pt}
\paragraph{Efficient Ensembles.}
Deep ensembles are widely used in classification challenges to boost the performance of CNNs~\cite{bian2017revisiting,simonyan2014very,Szegedy_2015_CVPR,he2016deep}.
More recent results indicate that deep ensembles of small models can be more efficient than single large models on image classification~\cite{kondratyuk2020ensembling,lobacheva2020power,NIPS2016_c51ce410, NIPS2016_20d135f0,furlanello2018born}, and we extend these findings to video classification.

\section{Mobile Video Networks (\ournets)} \label{sec:approach}

This section describes our progressive three-step approach to \ournets.
We first detail the design space to search for \ournets.
Then we define the stream buffer and explain how it reduces the networks' memory footprints, followed by the temporal ensembling to improve accuracy.

\subsection{Searching for \ournet} \label{sec:nas}
Following the practice of 2D mobile network search~\cite{tan2019mnasnet, tan2019efficientnet}, we start with the TuNAS framework~\cite{bender2020can}, which is a scalable implementation of one-shot NAS with weight sharing on a supernetwork of candidate models, and repurpose it for 3D CNNs for video recognition.
We use Kinetics 600~\cite{kay2017kinetics} as the video dataset to search over for all of our models, consisting of 10-second video sequences each at 25fps for a total of 250 frames.

\begin{table}[t]
    \footnotesize
    \newcommand{\blocks}[3]{
        \multirow{3}{*}{
            \(\left[
            \begin{array}{c}
                \text{1$\times$1$^\text{2}$, #2}\\[-.1em] 
                \text{1$\times$3$^\text{2}$, #2}\\[-.1em] 
                \text{1$\times$1$^\text{2}$, #1}
            \end{array}
            \right]\)$\times$#3
        }
    }
    \newcommand{\blocket}[4]{\multirow{3}{*}{\(\left[\begin{array}{c}\text{1$\times$1$^\text{2}$, #1}\\[-.1em] \text{$3$$\times$3$^\text{2}$, #2}\\[-.1em] \text{1$\times$1$^\text{2}$, #3}\end{array}\right]\)$\times$#4}
    }
    \newcommand{\blockt}[3]{\multirow{3}{*}{\(\left[\begin{array}{c}\text{\underline{3$\times$1$^\text{2}$}, #2}\\[-.1em] \text{1$\times$3$^\text{2}$, #2}\\[-.1em] \text{1$\times$1$^\text{2}$, #1}\end{array}\right]\)$\times$#3}
    }
    \newcommand{\blockseq}[3]{\text{#1$\times$#2$^\text{2}$, #3}\\[-.1em]}

    \begin{center}
 
    \begin{tabularx}{\columnwidth}{@{}Xcc@{}}
            \toprule
            \sc Stage & \sc Network Operations & \sc Output size \\
            \midrule
            data & stride $\tau$, RGB & $T\times S^\text{2}$ \\
            conv$_1$ & \multicolumn{1}{c}{$1\times k_1^\text{2}$, $c_1$}
            & $T\times \frac{S^\text{2}}{2}$ \\
            \midrule
            \multirow{2}{*}{block$_2$} & \multirow{2}{*}{
                \(\left[
                \begin{array}{c}
                    \blockseq{{$k_{2}^{\text{time}}$}}{{$(k_{2}^{\text{space}})$}}{{$c_2^\text{base}$, $c_2^\text{expand}$}}
                \end{array}
                \right]\)$\times L_2$} & 
                \multirow{2}{*}{$T\times \frac{S^\text{2}}{4}$}  \\
            & & \\
            & $\cdots$ & \\
            \multirow{2}{*}{block$_n$} & \multirow{2}{*}{
                \(\left[
                \begin{array}{c}
                    \blockseq{{$k_{n}^{\text{time}}$}}{{$(k_{n}^{\text{space}})$}}{{$c_n^\text{base}$, $c_n^\text{expand}$}}
                \end{array}
                \right]\)$\times L_n$} &
                \multirow{2}{*}{$T\times \frac{S^\text{2}}{2^n}$}  \\
            & & \\
            \midrule
            conv$_{n+1}$ & \multicolumn{1}{c}{$1\times1^\text{2}$, {$c_{n+1}^\text{base}$}}
            & $T\times \frac{S^\text{2}}{2^n}$ \\
            pool$_{n+2}$ & \multicolumn{1}{c}{$T\times \frac{S^\text{2}}{2^n}$}
            & $1\times1^\text{2}$ \\
            dense$_{n+3}$ & \multicolumn{1}{c}{$1\times1^\text{2}$, {$c_{n+3}^\text{base}$}}
            & $1\times1^\text{2}$ \\
            dense$_{n+4}$ & \multicolumn{1}{c}{$1\times1^\text{2}$, {\# classes}}
            & $1\times1^\text{2}$ \\
            \bottomrule
    \end{tabularx}

    \end{center}
    \vspace{-2pt}
    \caption{
        {\bf \ournet Search Space}. 
        Given an input video with $T$ frames and resolution $S^\text{2}$, at stage $i$ we search over base widths $c_i^{\text{base}}$ and the number of layers $L_i$ in the block.
        Within each layer we search for expansion widths $c_i^{\text{expand}}$, along with 3D convolutional kernel sizes $k_{i}^{\text{time}}\times(k_{i}^{\text{space}})^2 \in \{1, 3, 5, 7\} \times \{1, 3, 5, 7\}^2$.
    }
    \label{table:nas-ssd}
    \vspace{-5pt}
\end{table}

\vspace{-10pt}
\paragraph{\ournet Search Space.}
We build our base search space on MobileNetV3~\cite{howard2019searching}, which provides a strong baseline for mobile CPUs.
It consists of several blocks of inverted bottleneck layers with varying filter widths, bottleneck widths, block depths, and kernel sizes per layer. 
Similar to X3D~\cite{feichtenhofer2020x3d}, we expand the 2D blocks in MobileNetV3 to deal with 3D video input.
Table~\ref{table:nas-ssd} provides a basic overview of the search space, detailed as follows.

We denote by $T \times S^2 = 50 \times 224^2$ and $\tau = 5$ (5fps) the dimensions and frame stride, respectively, of the input to the target \ournets.
For each block in the network, we search over the base filter width $c^{\text{base}}$ and the number of layers $L \leq 10$ to repeat within the block. We apply multipliers $\{0.75, 1, 1.25\}$ over the feature map channels within every block, rounded to a multiple of 8. 
We set $n = 5$ blocks, with strided spatial downsampling for the first layer in each block except the 4th block (to ensure the last block has spatial resolution $7^2$).
The blocks progressively increase their feature map channels: $\{16, 24, 48, 96, 96, 192\}$.
The final convolution layer's base filter width is $512$, followed by a $2048$D dense layer before the classification layer.

With the new time dimension, we define the 3D kernel size within each layer, $k^{\text{time}}\times(k^{\text{space}})^2$, to be chosen as one of the following: \{1x3x3, 1x5x5, 1x7x7, 5x1x1, 7x1x1, 3x3x3, 5x3x3\} (we remove larger kernels from consideration).
These choices enable a layer to focus on and aggregate different dimensional representations, expanding the network's receptive field in the most pertinent directions while reducing FLOPs along other dimensions.
Some kernel sizes may benefit from having different numbers of input filters, so we search over a range of bottleneck widths $c^{\text{expand}}$ defined as multipliers in $\{1.5, 2.0, 2.5, 3.0, 3.5, 4.0\}$ relative to $c^{\text{base}}$.
Each layer surrounds the 3D convolution with two 1x1x1 convolutions to expand and project between $c^{\text{base}}$ and $c^{\text{expand}}$.
We do not apply any temporal downsampling to enable frame-wise prediction.

Instead of applying spatial squeeze-and-excitation (SE)~\cite{hu2018squeeze}, we use SE blocks to aggregate spatiotemporal features via 3D average pooling, applying it to every bottleneck block as in \cite{howard2019searching, tan2019efficientnet}.
We allow SE to be searchable, optionally disabling it to conserve FLOPs.

\vspace{-10pt}
\paragraph{Scaling the Search Space.}
Our base search space forms the basis for \ournet-A2.
For the other \ournets, we apply a compound scaling heuristic similar to the one used in EfficientNet~\cite{tan2019efficientnet}.
The major difference in our approach is that we scale the {\it search space} itself rather than a single model (i.e., search spaces for models A0-A5).
Instead of finding a good architecture and then scaling it, we search over all scalings of all architectures, broadening the range of possible models.

We use a small random search to find the scaling coefficients (with an initial target of 300 MFLOPs per frame), which roughly double or halve the expected size of a sampled model in the search space.
For the choice of coefficients, we resize the base resolution $S^2$, frame stride $\tau$, block filter width $c^{\text{base}}$, and block depths $L$.
We perform the search on different FLOPs targets to produce a family of models ranging from MobileNetV3-like sizes up to the sizes of ResNet3D-152~\cite{he2016deep, hara2018can}.
Appendix~\appendixsearchspace provides more details of the search space, the scaling technique, and a description of the search algorithm. 

The \ournet search space gives rise to a family of versatile networks, which outperform state-of-the-art efficient video recognition CNNs on popular benchmark datasets.
However, their memory footprints grow proportionally to the number of input frames, making them difficult to handle long videos on mobile devices.
The next subsection introduces a stream buffer to reduce the networks' memory consumption from linear to constant in video length.

\begin{figure*}
    \centering
    \includegraphics[width=\textwidth]{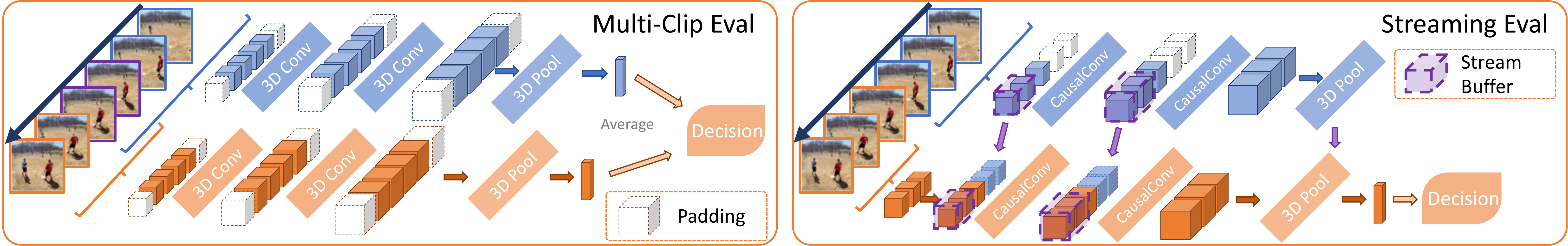}
    \caption{
        {\bf Streaming Evaluation vs. Multi-Clip Evaluation}.
        In multi-clip evaluation (a type of test-time data augmentation), we embed overlapping subclips of an input video with 3D convolutions and average the logits.
        In streaming evaluation, we use the stream buffer to carry forward input features between non-overlapping subclips and apply causal operations, thereby allowing the temporal receptive field to cover the whole video.
        This stream buffer increases accuracy while retaining the benefits of reduced memory from multi-clip evaluation.
    }
    \label{fig:multi-clip-vs-stream-buffer}
    \vspace{-10pt}
\end{figure*}

\subsection{The Stream Buffer with Causal Operations} \label{sec:stream-buffer}
Suppose we have an input video $\vec{x}$ with $T$ frames that may cause a model to exceed a set memory budget.
A common solution to reduce memory is  multi-clip evaluation~\cite{simonyan2014two, wang2015towards}, where the model averages predictions across $n$ overlapping subclips with $T^{\text{clip}} < T$ frames each, as seen in Figure~\ref{fig:multi-clip-vs-stream-buffer} (left).
It reduces memory consumption to $O(T^{\text{clip}})$.
However, it poses two major disadvantages: 
1) It limits the temporal receptive fields to each subclip and ignores long-range dependencies, potentially harming accuracy.
2) It recomputes frame activations which overlap, reducing efficiency.

\vspace{-10pt}
\paragraph{Stream Buffer.}
To overcome the above mentioned limitations, we propose stream buffer as a mechanism to cache feature activations on the boundaries of subclips, allowing us to expand the temporal receptive field across subclips and requiring no recomputation, as shown in Figure~\ref{fig:multi-clip-vs-stream-buffer} (right).

Formally, let $\vec{x}_{i}^{\text{clip}}$ be the current subclip (raw input or activation) at step $i < n$, where we split the video into $n$ adjacent \emph{non-overlapping} subclips  of length $T^\text{clip}$ each.
We start with a zero-initialized tensor representing our buffer $B$ with length $b$ along the time dimension and whose other dimensions match $\vec{x}_{i}^{\text{clip}}$. We compute the feature map $F_{i}$ of the buffer concatenated~($\oplus$) with the subclip along the time dimension as:
\begin{align}
    F_{i} = f(B_{i} \oplus \vec{x}_{i}^{\text{clip}})
\end{align}
where $f$ represents a spatiotemporal operation (e.g., 3D convolution).
When processing the next clip, we update the contents of the buffer to: 
\begin{align}
    B_{i+1} = (B_{i} \oplus \vec{x}_{i}^{\text{clip}})_{[-b:]}
\end{align}
where we denote $[-b:]$ as a selection of the last $b$ frames of the concatenated input.
As a result, our memory consumption is dependent on $O(b + T^{\text{clip}})$, which is constant as the total video frames $T$ or number of subclips $n$ increase.

\vspace{-10pt}
\paragraph{Relationship to TSM.}
The Temporal Shift Module (TSM)~\cite{lin2019tsm} can be seen as a special case of the stream buffer, where $b = 1$ and $f$ is an operation that shifts a proportion of channels in the buffer $B_t = \vec{x}_{t-1}$ to the input $\vec{x}_{t}$ before computing a spatial convolution at frame $t$.

\vspace{-2pt}
\subsubsection{Causal Operations}
A reasonable approach to fitting 3D CNNs' operations to the stream buffer is to enforce \emph{causality}, i.e., any features must not be computed  from future frames.
This has a number of advantages, including the ability to reduce a subclip $\vec{x}_i^{\text{clip}}$ down to a single frame without affecting activations or predictions, and enables 3D CNNs to work on streaming video for online inference.
While it is possible to use non-causal operations, e.g., buffering in both temporal directions, we would lose online modeling capabilities which is a desirable property for mobile.

\vspace{-10pt}
\paragraph{Causal Convolution (CausalConv).}
By leveraging the translation equivariant property of convolution, we replace all temporal convolutions with CausalConvs~\cite{oord2016wavenet}, effectively making them unidirectional along the temporal dimension.
Concretely, we first compute padding to balance the convolution across all axes and then move any padding after the final frame and merge it with any padding before the first frame.
See Appendix~\appendixmoreresults for an illustration of how the receptive field differs from standard convolution, as well as a description of the causal padding algorithm.

When using a stream buffer with CausalConv, we can replace causal padding with the buffer itself, carrying forward the last few frames from a previous subclip and copying them into the padding of the next subclip.
If we have a temporal kernel size of $k$ (and we do not use any strided sampling), then our padding and therefore buffer width becomes $b = k - 1$. 
Usually, $k = 3$ which implies $b = 2$, resulting in a small memory footprint.
Stream buffers are only required before layers that aggregate features across multiple frames, so spatial and pointwise convolutions (e.g., 1x3x3, 1x1x1) can be left as-is, further saving memory.

\vspace{-10pt}
\paragraph{Cumulative Global Average Pooling (CGAP).}
We use CGAP to approximate any global average pooling involving the temporal dimension.
For any activations up to frame $T'$, we can compute this as a cumulative sum:
\begin{align}
    \operatorname{CGAP}(\vec{x}, T') = \frac{1}{T'} \sum_{t=1}^{T'} \vec{x}_t,
\end{align}
where $\vec{x}$ represents a tensor of activations. 
To compute CGAP causally, we keep a single-frame stream buffer storing the cumulative sum up to $T'$.

\vspace{-10pt}
\paragraph{CausalSE with Positional Encoding.}
We denote CausalSE as the application of CGAP to SE, where we multiply the spatial feature map at frame $t$ with the SE computed from $\operatorname{CGAP}(\vec{x}, t)$.
From our empirical results, CausalSE is prone to instability likely due to the SE projection layers have a difficult time determining the quality of the CGAP estimate, which has high variance early in the video.
To combat this problem, we apply a sine-based fixed positional encoding (\textsc{PosEnc}) scheme inspired by Transformers~\cite{vaswani2017attention, liu2018intriguing}.
We directly use frame index as the position and sum the vector with CGAP output before applying the SE projection.

\vspace{-2pt}
\subsubsection{Training and Inference with Stream Buffers}

\vspace{-2pt}
\paragraph{Training.}
To reduce the memory requirements during training, we use a recurrent training strategy where we split a given batch of examples into $n$ subclips, applying a forward pass that outputs a prediction for each subclip, using stream buffers to cache activations.
However, we do not backpropagate gradients past the buffer so that the memory of previous subclips can be deallocated.
Instead, we compute losses and accumulate computed gradients between subclips, similar to batch gradient accumulation.
This allows the network to account for all $T = nT^{\text{clip}}$ frames, performing $n$ forward passes before applying the gradients.
This training strategy allows the network to learn longer term dependencies thus results in better accuracy than a model trained with shorter video length (see Appendix~\appendixmoreresults).

We can set $T^{\text{clip}}$ to any value without affecting accuracy.
However, ML accelerators (e.g., GPUs) benefit from multiplying large tensors, so for training we typically set a value of $T^{\text{clip}} \in \{8, 16, 32\}$.
This accelerates training while allowing careful control of memory cost.

\vspace{-10pt}
\paragraph{Online Inference.}
One major benefit of using causal operations like CausalConv and CausalSE is to allow a 3D video CNN to work online.
Similar to training, we use the stream buffer to cache activations between subclips.
However, we can set the subclip length to a single frame ($T^{\text{clip}} = 1$) for maximum memory savings.
This also reduces the latency between frames, enabling the model to output predictions frame-by-frame on a streaming video, accumulating new information incrementally like a recurrent network (RNN)~\cite{hochreiter1997long}.
But unlike traditional convolutional RNNs, we can input a variable number of frames per step to produce the same output.
For streaming architectures with CausalConv, we predict a video's label by pooling the frame-by-frame output features using CGAP.

\subsection{Temporal Ensembles}
The stream buffers can reduce \ournets' memory footprints up to an order of magnitude in the cost of about 1\% accuracy drop on Kinetics 600. 
We can restore this accuracy using a simple ensembling strategy.
We train two \ournets independently with the same architecture, but halve the frame-rate, keeping the temporal duration the same (resulting in half the input frames).
We input a video into both networks, with one network having frames offset by one frame and apply an arithmetic mean on the unweighted logits before applying softmax.
This method results in a two-model ensemble with the same FLOPs as a single model before halving the frame-rate, providing prediction with enriched representations.
In our observations, despite the fact that both models in the ensemble may have lower accuracy than the single model individually, together when ensembled they can have higher accuracy than the single model.

\section{Experiments on Video Classification} \label{sec:experiments}
In this section, we evaluate \ournets' accuracy, efficiency, and memory consumption during inference on five representative action recognition datasets.

\vspace{-10pt}
\paragraph{Datasets.}
We report results on all Kinetics datasets, including Kinetics 400~\cite{carreira2017quo, kay2017kinetics}, Kinetics 600~\cite{carreira2018short}, and Kinetics 700~\cite{carreira2019short}, which contain 10-second, 250-frame video sequences at 25 fps labeled with 400, 600, and 700 action classes, respectively.
We use examples that are available \emph{at the time of writing}, which is 87.5\%, 92.8\%, and 96.2\% of the training examples respectively (see Appendix~\appendixmoreresults).
Additionally, we experiment with Moments in Time~\cite{monfort2019moments}, containing 3-second, 75-frame sequences at 25fps in 339 action classes, and Charades~\cite{sigurdsson2016hollywood}, which has variable-length videos with 157 action classes where a video can contain multiple class annotations.
We include Something-Something V2~\cite{goyal2017something} and Epic Kitchens 100~\cite{damen2020rescaling} results in Appendix~\appendixmoreresults.

\vspace{-10pt}
\paragraph{Implementation Details.} 
For each dataset, all models are trained with RGB frames from scratch, i.e., we do not apply any pretraining.
For all datasets, we train with 64 frames (except when the inference frames are fewer) at various frame-rates, and run inference with the same frame-rate.

We run TuNAS using Kinetics 600 and keep 7 \ournets each having a FLOPs target used in~\cite{feichtenhofer2020x3d}.
As our models get larger, our scaling coefficients increase the input resolution, number of frames, depth, and feature width of the networks.
We also experiment with AutoAugment~\cite{cubuk2018autoaugment} augmentation used in image classification, i.e., we sample a random image augmentation for each video and apply the same augmentation for each frame.
For the architectures of the 7 models as well as training hyperparameters, see Appendix~\appendixarchitectures.

\vspace{-10pt}
\paragraph{Single-Clip vs. Multi-Clip Evaluation.}
We evaluate all our models with a single clip sampled from input video with a fixed temporal stride, covering the entire video duration.
When the single-clip and multi-clip evaluations use the same number of frames in total so that FLOPs are equivalent, we find that single-clip evaluation yields higher accuracy (see Appendix~\appendixmoreresults).
This can be due in part to 3D CNNs being able to model longer-range dependencies, even when evaluating on many more frames than it was trained on.
Since existing models commonly use multi-clip evaluation, we report the total FLOPs per video, not per clip, for a fair comparison.

However, single-clip evaluation can greatly inflate a network's peak memory usage (as seen in Figure~\ref{fig:k600-comparison}), which is likely why multi-clip evaluation is commonly used in previous work.
The stream buffer eliminates this problem, allowing \ournets to predict like they are embedding the full video, and incurs less peak memory than multi-clip evaluation.

We also reproduce X3D~\cite{feichtenhofer2020x3d}, arguably the most related work to ours, to test its performance under single-clip and 10-clip evaluation to provide more insights.
We denote 30-clip to be the evaluation strategy with 10 clips times three spatial crops for each video, while 10-clip just uses one spatial crop.
We avoid any spatial augmentation or temporal sampling in \ournets during inference to improve efficiency.

\vspace{-3pt}
\subsection{Comparison Results on Kinetics 600}
\vspace{-3pt}
\paragraph{\ournets without Stream Buffers.} 
Table~\ref{table:k600-comparison} presents the main results of seven \ournets on Kinetics 600 \emph{before applying the stream buffer}, mainly compared with various X3D models~\cite{feichtenhofer2020x3d}, which are recently developed for efficient video recognition. The columns of the table correspond to the Top-1 classification accuracy; GFLOPs per video a model incurs; resolution of the input video frame (where we shorten $224^2$ to 224); input frames per video, where $30\times4$ means the 30-clip evaluation with 4 frames as input in each run; frames per second (FPS), determined by the temporal stride $\tau$ in the search space for \ournets; and a network's number of parameters.

\begin{table}[tbp]
    \newcommand{\frameinput}[2]{#1$\times$#2$^2$}
    \begin{center}
    \resizebox{\columnwidth}{!}{%
    \begin{tabular}{@{}lrrrrrr@{}}
    \toprule
        \sc Model & \sc Top-1 & \sc gflops & \sc Res & \sc Frames & \sc FPS & \sc Param \\
    \midrule
        \bf \ournet-A0 & \bf 71.5 & \bf 2.71 & 172 & 1$\times$50 & 5 & 3.1M \\
        MobileNetV3-S* \cite{howard2019searching} & 61.3 & 2.80 & 224 & 1$\times$50 & 5 & 2.5M \\
        MobileNetV3-S+TSM* \cite{lin2019tsm} & 65.5 & 2.80 & 224 & 1$\times$50 & 5 & 2.5M \\
        X3D-XS* \cite{feichtenhofer2020x3d} & 70.2 & 3.88 & 182 & 1$\times$20 & 2 & 3.8M \\
    \midrule
        \bf \ournet-A1 & \bf 76.0 & \bf 6.02 & 172 & 1$\times$50 & 5 & 4.6M \\
        X3D-S* \cite{feichtenhofer2020x3d} & 73.4 & 7.80 & 182 & 1$\times$40 & 4 & 3.8M \\
        X3D-S* \cite{feichtenhofer2020x3d} & 74.3 & 9.75 & 182 & 1$\times$50 & 5 & 3.8M \\
    \midrule
        \bf \ournet-A2 & \bf 77.5 & \bf 10.3 & 224 & 1$\times$50 & 5 & 4.8M \\
        MobileNetV3-L* \cite{howard2019searching} & 68.1 & 11.0 & 224 & 1$\times$50 & 5 & 5.4M \\
        MobileNetV3-L+TSM*~\cite{lin2019tsm} & 71.4 & 11.0 & 224 & 1$\times$50 & 5 & 5.4M  \\
        X3D-XS \cite{feichtenhofer2020x3d} & 72.3 & 23.3 & 182 & 30$\times$4 & 2 & 3.8M \\
        X3D-M* \cite{feichtenhofer2020x3d} & 76.9 & 19.4 & 256 & 1$\times$50 & 5 & 3.8M \\
    \midrule
        \bf \ournet-A3 & 80.8 & \bf 56.9 & 256 & 1$\times$120 & 12 & 5.3M \\
        \bf \ournet-A3 + AutoAugment & \bf 81.3 & \bf 56.9 & 256 & 1$\times$120 & 12 & 5.3M \\
        X3D-S \cite{feichtenhofer2020x3d} & 76.4 & 76.1 & 182 & 30$\times$13 & 4 & 3.8M \\
        X3D-L* \cite{feichtenhofer2020x3d} & 79.1 & 77.5 & 356 & 1$\times$50 & 5 & 6.1M \\
    \midrule
        \bf \ournet-A4 & 81.2 & \bf 105 & 290 & 1$\times$80 & 8 & 4.9M \\
        \bf \ournet-A4 + AutoAugment & \bf 83.0 & \bf 105 & 290 & 1$\times$80 & 8 & 4.9M \\
        X3D-M \cite{feichtenhofer2020x3d} & 78.8 & 186 & 256 & 30$\times$16 & 5 & 3.8M \\
        X3D-L* \cite{feichtenhofer2020x3d} & 80.7 & 187 & 356 & 1$\times$120 & 2 & 6.1M \\
        X3D-XL* \cite{feichtenhofer2020x3d} & 80.3 & 151 & 356 & 1$\times$50 & 5 & 11.0M \\
        I3D \cite{carreira2018short} & 71.6 & 216 & 224 & 1$\times$250 & 25 & 12M \\
        ResNet3D-50* & 78.7 & 390 & 224 & 1$\times$250 & 25 & 34.0M \\
    \midrule
        \bf \ournet-A5 & 82.7 & \bf 281 & 320 & 1$\times$120 & 12 & 15.7M \\
        \bf \ournet-A5 + AutoAugment & \bf 84.3 & \bf 281 & 320 & 1$\times$120 & 12 & 15.7M \\
        X3D-L \cite{feichtenhofer2020x3d} & 80.5 & 744 & 356 & 30$\times$16 & 5 & 6.1M \\
    \midrule
        \bf \ournet-A6 & 83.5 & \bf 386 & 320 & 1$\times$120 & 12 & 31.4M \\
        \bf \ournet-A6 + AutoAugment & \bf 84.8 & \bf 386 & 320 & 1$\times$120 & 12 & 31.4M \\
        ViViT-L/16x2~\cite{arnab2021vivit} & 83.0 & 3990 & 320 & 12$\times$32 & 12 & 88.9M \\
        TimeSformer-HR \cite{bertasius2021space} & 82.4  & 5110 & 224 & 3$\times$8 & 1.5 & 120M \\
        X3D-XL \cite{feichtenhofer2020x3d} & 81.9 & 1452 & 356 & 10$\times$16 & 5 & 11.0M \\
        ResNet3D-152* & 81.1 & 1400 & 224 & 1$\times$250 & 25 & 80.1M \\
        ResNet3D-50-G~\cite{li2020perf} & 82.0 & 3666 & 224 & 1$\times$250 & 25 & - \\
        SlowFast-R50 \cite{feichtenhofer2019slowfast} & 78.8 & 1080 & 256 & 30$\times$16 & 5 & 34.4M \\
        SlowFast-R101 \cite{feichtenhofer2019slowfast} & 81.8 & 7020 & 256 & 30$\times$16 & 5 & 59.9M \\
        LGD-R101 \cite{qiu2019learning} & 81.5 & - & 224 & 15$\times$16 & 25 & - \\
    \bottomrule
    \end{tabular}
    }
    \end{center}
    \caption{
        {\bf Accuracy of \ournet on Kinetics 600}.
        We measure total GFLOPs per video across all frames, and report the inference resolution (res),  number of clips $\times$ frames per clip (frames), and frame rate (fps) of each video clip.
        * denotes our reproduced models.
        For X3D, we report \emph{inference} resolution, which differs from training.
        We report all datapoints to the best knowledge available.
    }
    \label{table:k600-comparison}
    \vspace{-5pt}
\end{table}




\ournet-A0 has fewer GFLOPs and is 10\% more accurate than the frame-based MobileNetV3-S~\cite{howard2019searching} (where we train MobileNetV3 using our training setup, averaging logits across frames). 
\ournet-A0 also outperforms X3D-S in terms of both accuracy and GFLOPs. 
\ournet-A1 matches the GFLOPs of X3D-S, but its accuracy is 2\% higher than X3D-S.


Growing the target GFLOPs to the range between X3D-S and 30-clip X3D-XS, we arrive at \ournet-A2. We can achieve a little higher accuracy than 30-clip X3D-XS or X3D-M by using almost half of their GFLOPs.
Additionally, we include the frame-by-frame MobileNetV3-L and verify that it can benefit from TSM~\cite{lin2019tsm} by about 3\%.

There are more significant margins between larger \ournets (A3--A6) and their counterparts in the X3D family. 
It is not surprising because NAS should intuitively be more advantageous over the handcrafting method for X3D when the design space is large.  
\ournet-A5 and \ournet-A6 outperform several state-of-the-art video networks, including recent Transformer models like ViViT~\cite{arnab2021vivit} and TimeSformer~\cite{bertasius2021space} (see the last 6 rows of Table~\ref{table:k600-comparison}). 
\ournet-A6 with AutoAugment achieves 84.8\% accuracy (without pretraining) while still being substantially more efficient than comparable models (most often by an order of magnitude).
Even when compared to fully Transformer~\cite{vaswani2017attention} models like TimeSformer-HR~\cite{bertasius2021space}, \ournet-A6 outperforms it by 1\% accuracy and using 40\% of the FLOPs.


\begin{table}[tbp]
    \begin{center}
    \resizebox{\columnwidth}{!}{%
    \begin{tabular}{@{}lrrrrrr@{}}
    \toprule
        \sc Model & \sc Top-1 & \sc Res & \sc Frames & \sc FPS & \sc gflops & \sc Mem (MB) \\
    \midrule
        MobileNetV3-L* \cite{howard2019searching} & \bf 68.1 & 224 & 1$\times$50 & 5 & 11.0 & \bf 23 \\
    \midrule
        \ournet-A0 & \bf 71.5 & 172 & 1$\times$50 & 5 & 2.71 & 173 \\
        \ournet-A0-Stream & 70.3 & 172 & 1$\times$50 & 5 & 2.73 & \bf 71 \\
    \midrule
        \ournet-A1 & \bf 76.0 & 172 & 1$\times$50 & 5 & 6.02 & 191 \\
        \ournet-A1-Stream & 75.6 & 172 & 1$\times$50 & 5 & 6.06 & \bf 72 \\
        \ournet-A1-Stream-Ens (x2) & 75.9 & 172 & 1$\times$25 & 2.5 & 6.06 & \bf 72 \\
    \midrule
        \ournet-A2 & \bf 77.5 & 224 & 1$\times$50 & 5 & 10.3 & 470 \\
        \ournet-A2-Stream & 76.5 & 224 & 1$\times$50 & 5 & 10.4 & \bf 85 \\
        \ournet-A2-Stream-Ens (x2) & 77.0 & 224 & 1$\times$25 & 2.5 & 10.4 & \bf 85 \\
    \midrule
        \ournet-A3 & 80.8 & 256 & 1$\times$120 & 12 & 56.9 & 1310 \\
        \ournet-A3-Stream & 79.6 & 256 & 1$\times$120 & 12 & 57.1 & \bf 82 \\
        \ournet-A3-Stream-Ens (x2) & \bf 80.4 & 256 & 1$\times$60 & 6 & 57.1 & \bf 82 \\
    \midrule
        \ournet-A4 & 81.2 & 290 & 1$\times$80 & 8 & 105 & 1390 \\
        \ournet-A4-Stream & 80.5 & 290 & 1$\times$80 & 8 & 106 & \bf 112 \\
        \ournet-A4-Stream-Ens (x2) & \bf 81.4 & 290 & 1$\times$40 & 4 & 106 & \bf 112 \\
    \midrule
        \ournet-A5 & 82.7 & 320 & 1$\times$120 & 12 & 281 & 2040 \\
        \ournet-A5-Stream & 82.0 & 320 & 1$\times$120 & 12 & 282 & \bf 171 \\
        \ournet-A5-Stream-Ens (x2) & \bf 82.9 & 320 & 1$\times$60 & 6 & 282 & \bf 171 \\
    \midrule
        ResNet3D-50 & \bf 78.7 & 224 & 1$\times$250 & 25 & 390 & 3040 \\
        ResNet3D-50-Stream & 76.9 & 224 & 1$\times$250 & 25 & 390 & \bf 2600 \\
        ResNet3D-50-Stream-Ens (x2) & 78.6 & 224 & 1$\times$125 & 12.5 & 390 & \bf 2600 \\
    \bottomrule
    \end{tabular}
    }
    \end{center}
    \vspace{-2pt}
    \caption{
        {\bf Base vs. Streaming Architectures} on Kinetics 600.
        We and report the inference resolution (res), number of clips $\times$ frames per clip (frames), and frame rate (fps) for each video.
        We measure the total GFLOPs per video across all frames.
        We denote ``Stream'' to be causal models using a stream buffer frame-by-frame, and ``Ens'' to be two ensembled models (with half the input frames so FLOPs are equivalent).
        Memory usage is measured in peak MB for a single video clip.
        * denotes our reproduced models.
    }
    \label{table:balanced-vs-streaming}
    \vspace{-10pt}
\end{table}

\begin{figure}[t]
    \centering
    \includegraphics[width=1.0\linewidth]{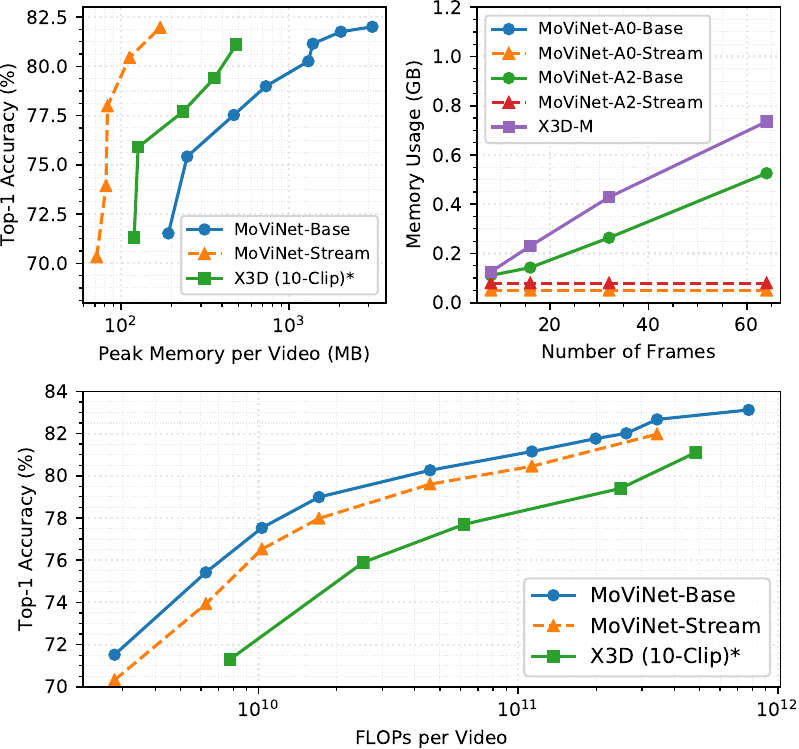}
    \caption{
        {\bf Effect of Streaming \ournets on Memory} on Kinetics 600.
        Top left: comparison of accuracy vs. max memory usage on a V100 GPU on our models, progressively increasing in size.
        We evaluate two versions of \ournet: a base version without a stream buffer and a causal version with a stream buffer.
        Note that memory may be inflated due to padding and runtime overhead.
        Top right: comparison of max memory usage on a V100 GPU as a function of the number of input frames.
        Bottom: the classification accuracy.
        * denotes our reproduced models.
    }
    \label{fig:frame-stream}
    \vspace{-10pt}
\end{figure}

\vspace{-10pt}
\paragraph{\ournets with Stream Buffers.}
Our base \ournet architectures may consume lots of memory in the absence of modifications, especially as the model sizes and input frames grow.
Using the stream buffer with causal operations, we can have an order of magnitude peak memory reduction for large networks (\ournets A3-A6), as shown in the last column of Table~\ref{table:balanced-vs-streaming}.

Moreover, Figure~\ref{fig:frame-stream} visualizes the streaming architectures' effect on memory. 
From the left panel at the top, we see that our \ournets are more accurate and more memory-efficient across all model sizes compared to X3D, which employs multi-clip evaluation.  
We also demonstrate constant memory as we scale the total number of frames in the input receptive field at the top's right panel. 
The bottom panel indicates that the streaming \ournets remain efficient in terms of the GFLOPs per input video.

We also apply our stream buffer to ResNet3D-50 (see the last two rows in Table~\ref{table:balanced-vs-streaming}).
However, we do not see as much of a memory reduction, likely due to larger overhead when using full 3D convolution as opposed to the depthwise convolution in \ournets.

\vspace{-10pt}
\paragraph{\ournets with Stream Buffers and Ensembling.}
We see from Table~\ref{table:balanced-vs-streaming} only a small 1\% accuracy drop across all models after applying the stream buffer. 
We can restore the accuracy using the temporal ensembling without any additional inference cost.
Table~\ref{table:balanced-vs-streaming} reports the effect of ensembling two models trained at half the frame rate of the original model (so that GFLOPs remain the same). 
We can see the accuracy improvements in all streaming architectures, showing that ensembling can bridge the gap between streaming and non-streaming architectures, especially as model sizes grow.
It is worth noting that, unlike prior works, the ensembling balances accuracy and efficiency (GFLOPs) in the same spirit as~\cite{kondratyuk2020ensembling}, not just to boost the accuracy.

\begin{figure*}[tbp]
    \begin{center}
    \includegraphics[width=1.0\linewidth]{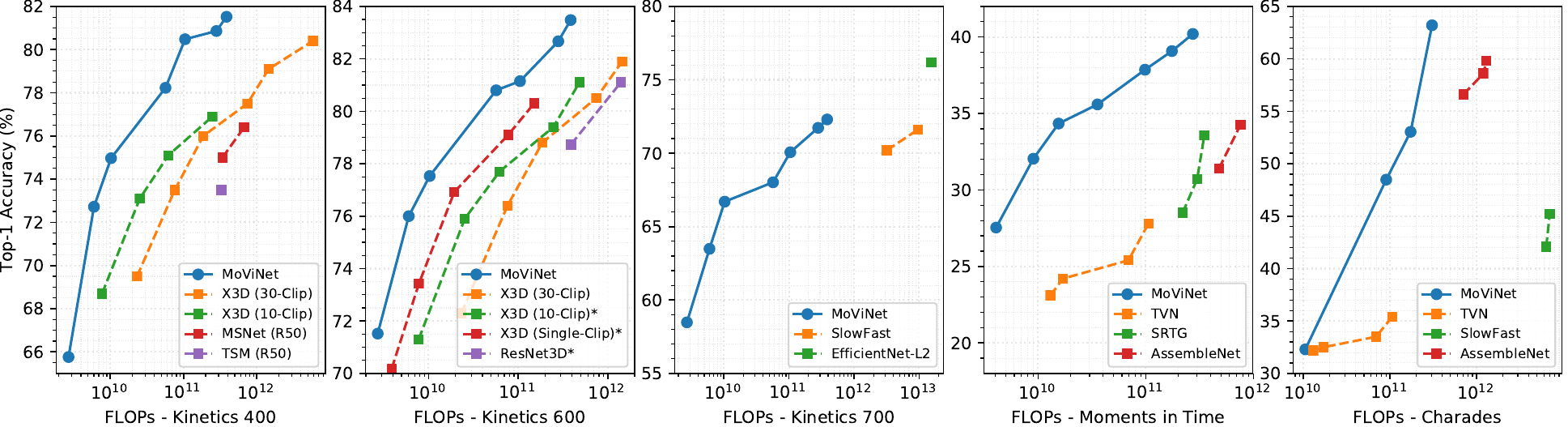}
    \end{center}
    \vspace{-8pt}
    \caption{
        {\bf Accuracy vs. FLOPs Comparison} across 5 large-scale action recognition datasets.
        Each series represents a model family, with points representing individual models ordered by FLOPs.
        We report FLOPs per video using single-clip evaluation for all \ournets and compare with competitive multi-clip (and reproduced single-clip) models, using a log-scale on the x-axis.
        * denotes reproduced models.
        For Charades, we evaluate on MoViNet A1, A4, A5, and A6 only.
    }
    \label{fig:all-flops-comparison}
    \vspace{-10pt}
\end{figure*}

\subsection{Comparison Results on Other Datasets}

Figure~\ref{fig:all-flops-comparison} summarizes the main results of \ournets on all the five datasets along with state-of-the-art models that have results reported on the respective datasets. We compare \ournets with X3D~\cite{feichtenhofer2020x3d}, MSNet~\cite{kwon2020motionsqueeze}, TSM~\cite{lin2019tsm}, ResNet3D~\cite{he2016deep}, SlowFast~\cite{feichtenhofer2019slowfast}, EfficientNet-L2~\cite{xie2020self}, TVN~\cite{piergiovanni2020tiny}, SRTG~\cite{stergiou2020learn}, and AssembleNet~\cite{ryoo2019assemblenet, ryoo2020assemblenetpp}.
Appendix~\appendixmoreresults tabulates the results with more details.

Despite only searching for efficient architectures on Kinetics 600, NAS yields models that drastically improve over prior work on other datasets as well.
On Moments in Time, our models are 5-8\% more accurate than Tiny Video Networks (TVNs)~\cite{piergiovanni2020tiny} at low GFLOPs, and \ournet-A5 achieves 39.9\% accuracy, outperforming AssembleNet~\cite{ryoo2019assemblenet} (34.3\%) which uses optical flow as additional input (while our models do not).
On Charades, \ournet-A5 achieves the accuracy of 63.2\%, beating AssembleNet++~\cite{ryoo2020assemblenetpp} (59.8\%) which uses optical flow and object segmentation as additional inputs.
Results on Charades provide evidence that our models are also capable of sophisticated temporal understanding, as these videos can have longer duration clips than what is seen in Kinetics and Moments in Time.

\subsection{Additional Analyses}

\paragraph{\ournet Operations.} We provide some ablation studies about
some critical \ournet operations in Table~\ref{table:stream-ablation}.
For the base network without the stream buffer, SE is vital for achieving high accuracy; \ournet-A1's accuracy drops by 2.9\% if we remove SE. 
We see a much larger accuracy drop when using CausalConv without SE than CausalConv with a global SE, which indicates that the global SE can take some of the role of standard Conv to extract information from future frames.
However, when we switch to a fully streaming architecture with CausalConv and CausalSE, this information from future frames is no longer available, and we see a large drop in accuracy, but still significantly improved from CausalConv without SE.
Using PosEnc, we can gain back some accuracy in the causal model.

\begin{table}[tbp]
    \centering
    \newcommand{\ck}{\checkmark}
    \begin{center}
    \resizebox{0.9\columnwidth}{!}{%
    \begin{tabular}{@{}lcccccc@{}}
    \toprule
       \sc Model & \sc CausalConv & \sc SE & \sc CausalSE & \sc PosEnc & \sc Top-1 & \sc gflops \\
    \midrule
                   &     &     &     &     & 73.3 & \bf 6.04 \\
                   & \ck &     &     &     & 72.1 & \bf 6.04 \\
                   & \ck &     & \ck &     & 73.5 & 6.06 \\
        \ournet-A1 & \ck &     & \ck & \ck & 74.0 & 6.06 \\
                   & \ck & \ck &     &     & 74.9 & 6.06 \\
                   &     & \ck &     &     & \bf 75.2 & 6.06 \\
    \midrule
                   &     &     &     &     & 77.7 & \bf 56.9 \\
        \ournet-A3 & \ck &     & \ck &     & 79.0 & 57.1 \\
                   & \ck &     & \ck & \ck & 79.6 & 57.1 \\
                   &     & \ck &     &     & \bf 80.3  & 57.1 \\
    \bottomrule
    \end{tabular}
    }
    \end{center}
    \vspace{-2pt}
    \caption{
        {\bf \ournet Operations Ablation} on Kinetics 600.
        We compare different configurations on \ournet-A1, including Conv/CausalConv, SE/CausalSE/No SE, and PosEnc, and report accuracy and GFLOPs per video.
    }
    \label{table:stream-ablation}
    \vspace{-10pt}
\end{table}

\begin{table}[t]
    \newcommand{\blocks}[3]{
        \multirow{3}{*}{
            \(\left[
            \begin{array}{c}
                \text{1$\times$1$^\text{2}$, #2}\\[-.1em] 
                \text{1$\times$3$^\text{2}$, #2}\\[-.1em] 
                \text{1$\times$1$^\text{2}$, #1}
            \end{array}
            \right]\)$\times$#3
        }
    }
    \newcommand{\blocket}[4]{\multirow{3}{*}{\(\left[\begin{array}{c}\text{1$\times$1$^\text{2}$, #1}\\[-.1em] \text{$3$$\times$3$^\text{2}$, #2}\\[-.1em] \text{1$\times$1$^\text{2}$, #3}\end{array}\right]\)$\times$#4}
    }
    \newcommand{\blockt}[3]{\multirow{3}{*}{\(\left[\begin{array}{c}\text{\underline{3$\times$1$^\text{2}$}, #2}\\[-.1em] \text{1$\times$3$^\text{2}$, #2}\\[-.1em] \text{1$\times$1$^\text{2}$, #1}\end{array}\right]\)$\times$#3}
    }
    \newcommand{\blockseq}[3]{\text{#1$\times$#2$^\text{2}$, #3}\\[-.1em]}

    \begin{center}
    \resizebox{0.7\columnwidth}{!}{%
    \begin{tabularx}{1.0 \columnwidth}{@{}Xcl@{}}
        \toprule
        \sc Stage & \sc Operation & \sc Output size \\
        \midrule
        data & stride 5, RGB & $50\times224^\text{2}$ \\
        conv$_1$ & \multicolumn{1}{c}{$1\times3^\text{2}$, {16}}
        & $50\times112^\text{2}$ \\
        \midrule
        block$_2$ & \multirow{3}{*}{
            \(\left[
            \begin{array}{c}
                \blockseq{{1}}{{5}}{{16, 40}}
                \blockseq{{3}}{{3}}{{16, 40}}
                \blockseq{{3}}{{3}}{{16, 64}}
            \end{array}
            \right]\)} & 
            \multirow{3}{*}{$50\times56^\text{2}$}  \\
        &  & \\
        &  & \\
        block$_3$ & \multirow{5}{*}{
            \(\left[
            \begin{array}{c}
                \blockseq{{3}}{{3}}{{40, 96}}
                \blockseq{{3}}{{3}}{{40, 120}}
                \blockseq{{3}}{{3}}{{40, 96}}
                \blockseq{{3}}{{3}}{{40, 96}}
                \blockseq{{3}}{{3}}{{40, 120}}
            \end{array}
            \right]\)} & 
            \multirow{5}{*}{$50\times28^\text{2}$}  \\
        &  & \\
        &  & \\
        &  & \\
        &  & \\
        block$_4$ & \multirow{5}{*}{
            \(\left[
            \begin{array}{c}
                \blockseq{{5}}{{3}}{{72, 240}}
                \blockseq{{3}}{{3}}{{72, 155}}
                \blockseq{{3}}{{3}}{{72, 240}}
                \blockseq{{3}}{{3}}{{72, 192}}
                \blockseq{{3}}{{3}}{{72, 240}}
            \end{array}
            \right]\)} & 
            \multirow{5}{*}{$50\times14^\text{2}$}  \\
        &  & \\
        &  & \\
        &  & \\
        &  & \\
        block$_5$ & \multirow{6}{*}{
            \(\left[
            \begin{array}{c}
                \blockseq{{5}}{{3}}{{72, 240}}
                \blockseq{{3}}{{3}}{{72, 240}}
                \blockseq{{3}}{{3}}{{72, 240}}
                \blockseq{{3}}{{3}}{{72, 240}}
                \blockseq{{1}}{{5}}{{72, 144}}
                \blockseq{{3}}{{3}}{{72, 240}}
            \end{array}
            \right]\)} & 
            \multirow{6}{*}{$50\times14^\text{2}$}  \\
        &  & \\
        &  & \\
        &  & \\
        &  & \\
        &  & \\
        block$_6$ & \multirow{7}{*}{
            \(\left[
            \begin{array}{c}
                \blockseq{{5}}{{3}}{{144, 480}}
                \blockseq{{1}}{{5}}{{144, 384}}
                \blockseq{{1}}{{5}}{{144, 384}}
                \blockseq{{1}}{{5}}{{144, 480}}
                \blockseq{{1}}{{5}}{{144, 480}}
                \blockseq{{3}}{{3}}{{144, 480}}
                \blockseq{{1}}{{3}}{{144, 576}}
            \end{array}
            \right]\)} & 
            \multirow{7}{*}{$50\times7^\text{2}$}  \\
        &  & \\
        &  & \\
        &  & \\
        &  & \\
        &  & \\
        &  & \\
        \midrule
        conv$_7$ & \multicolumn{1}{c}{$1\times1^\text{2}$, {640}}
        & $50\times7^\text{2}$ \\
        pool$_8$ & \multicolumn{1}{c}{$50\times7^\text{2}$}
        & $1\times1^\text{2}$ \\
        dense$_9$ & \multicolumn{1}{c}{$1\times1^\text{2}$, {2048}}
        & $1\times1^\text{2}$ \\
        dense$_{10}$ & \multicolumn{1}{c}{$1\times1^\text{2}$, {600}}
        & $1\times1^\text{2}$ \\
        \bottomrule
    \end{tabularx}
    }
    \end{center}
    \vspace{-5pt}
    \caption{
        {\bf \ournet-A2 Architecture} searched by TuNAS, running 50 frames on Kinetics 600.
        See Table~\ref{table:nas-ssd} for the search space definition detailing the meaning of each component.
    }
    \label{table:a2-architecture}
    \vspace{-10pt}
\end{table}

\vspace{-10pt}
\paragraph{\ournet Architectures.}
We provide the architecture description of \ournet-A2 in Table~\ref{table:a2-architecture} --- Appendix~\appendixarchitectures has the detailed architectures of other \ournets.
Most notably, the network prefers large bottleneck width multipliers in the range [2.5, 3.5], often expanding or shrinking them after each layer.
In contrast, X3D-M with similar compute requirements has a wider base feature width with a smaller constant bottleneck multiplier of 2.25.
The searched network prefers balanced 3x3x3 kernels, except at the first downsampling layers in the later blocks, which have 5x3x3 kernels.
The final stage almost exclusively uses spatial kernels of size 1x5x5, indicating that high-level features for classification benefit from mostly spatial features.
This comes at a contrast to S3D~\cite{xie2018rethinking}, which reports improved efficiency when using 2D convolutions at lower layers and 3D convolutions at higher layers.


\vspace{-10pt}
\subsubsection{Hardware Benchmark}

\begin{figure}[t]
    \centering
    \includegraphics[width=1.0\linewidth]{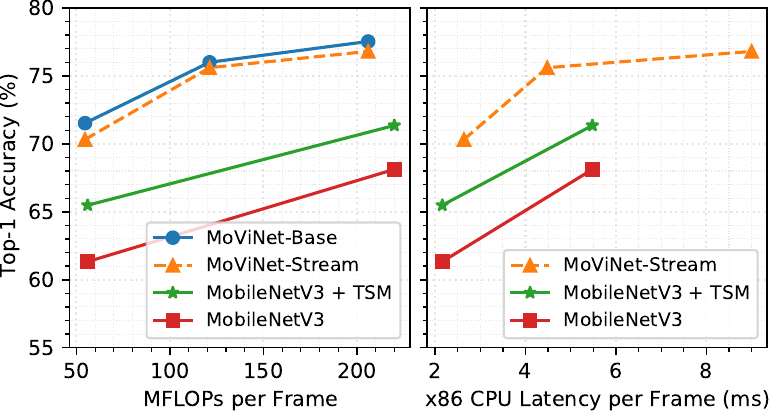}
    \caption{
        {\bf CPU Latency Comparison} on Kinetics 600.
        We compare the efficiency of \ournets  (A0, A1, A2) vs. MobileNetV3~\cite{howard2019searching} using FLOPs and benchmarked latency on an x86 Xeon CPU at 3.70GHz.
    }
    \label{fig:x86-comparison}
\end{figure}

\ournets A0, A1, and A2 represent the fastest models that would most realistically be used on mobile devices.
We compare them with MobileNetV3 in Figure~\ref{fig:x86-comparison} with respect to both FLOPs and real-time latency on an x86 Intel Xeon W-2135 CPU at 3.70GHz.
These models are comparable in per-frame computation cost, as we evaluate on 50 frames for all models.
From these results we can conclude that streaming \ournets can run faster on CPU while being more accurate at the same time, even with temporal modifications like TSM.
While there is a discrepancy between FLOPs and latency, searching over a latency target explicitly in NAS can reduce this effect.
However, we still see that FLOPs is a reasonable proxy metric for CPU latency, which would translate well for mobile devices.

\begin{table}[t]
    \begin{center}
    \newcommand{\ck}{\checkmark}
    \newcommand{\xmark}{\ding{55}}
    \resizebox{250pt}{!}{%
    \begin{tabular}{@{}lcrrr@{}}
    \toprule
        \sc Model & \sc Streaming & \sc Video (ms) & \sc Frame (ms) & \sc Top-1 \\
    \midrule
        MoViNet-A0-Stream & \ck & 183 & 3.7 & 70.3 \\
        MoViNet-A1-Stream & \ck & 315 & 6.3 & 75.6 \\
        MoViNet-A2-Stream & \ck & 325 & 6.5 & 76.5 \\
        MoViNet-A3-Stream & \ck & 1110 & 9.2 & 79.6 \\
        MoViNet-A4-Stream & \ck & 1130 & 14.1 & 80.5 \\
        MoViNet-A5-Stream & \ck & 2310 & 19.2 & 82.0 \\
    \midrule
        MobileNetV3-S* & \ck & 68 & 1.4 & 61.3 \\
        MobileNetV3-L* & \ck & 81 & 1.6 & 68.1 \\
    \midrule
        X3D-M* (Single Clip) & \xmark  & 345 & 6.9 & 76.9 \\
        X3D-XL* (Single Clip) & \xmark & 943 & 18.9 & 80.3 \\
    \bottomrule
    \end{tabular}
    }
    \end{center}
    \caption{
        {\bf Runtime on an Nvidia V100 GPU on Kinetics 600}. Latency is given for the entire video clip and per frame in ms. * denotes reproduced models.
    }
    \label{table:runtime-v100-comparison}
    \vspace{-10pt}
\end{table}

We also show benchmarks for \ournets running on an Nvidia V100 GPU in Table~\ref{table:runtime-v100-comparison}.
Similar to mobile CPU, our streaming model latency is comparable to single-clip X3D models.
However, we do note that MobileNetV3 can run faster than our networks on GPU, showing that the FLOPs metric for NAS has its limitations.
\ournets can be made more efficient by targeting real hardware instead of FLOPs, which we leave for future work.


\vspace{-2pt}
\section{Conclusion} \label{sec:conclusion}
\vspace{-2pt}
\ournets provide a highly efficient set of models that transfer well across different video recognition datasets.
Coupled with stream buffers, \ournets significantly reduce training and inference memory cost while also supporting online inference on streaming video.
We hope our approach to designing \ournets can provide improvements to future and existing models, reducing memory and computation costs in the process. 

{\small
\bibliographystyle{ieee_fullname}
\bibliography{efficient_video}
}

\newpage
\section*{Appendices} \label{sec:appendix}
\addcontentsline{toc}{section}{Appendix}
\renewcommand{\thesubsection}{\Alph{subsection}}

We supplement the main text by the following materials. 
\begin{itemize}
    \item {\bf Appendix~\ref{appendix:search-space}} provides more details of the search space, the technique to scale the search space, and the search algorithm.
    \item {\bf Appendix~\ref{appendix:architectures}} is about the neural architectures of \ournets A0-A7. 
    \item {\bf Appendix~\ref{appendix:more-results}} reports additional results on the datasets studied in the main text along with ablation studies.
\end{itemize}


\subsection{\ournet Architecture Search} \label{appendix:search-space}




\subsubsection{Scaling Algorithm for the Search Space}

To produce models that scale well, we progressively expand the search space across width, depth, input resolution, and frame rate, like EfficientNet~\cite{tan2019efficientnet}.
Specifically, we use a single scaling parameter $\phi$ to define the size of our search space.
Then we define the following coefficients:
\begin{align*}
    \text{depth: } d = \alpha^\phi = 1.36^\phi \\
    \text{base width: } w = \beta^\phi = 1.18^\phi \\
    \text{resolution: } r = \gamma^\phi = 1.16^\phi \\
    \text{frame-rate: } f = \delta^\phi = 1.24^\phi
\end{align*}
such that $\alpha \beta^2 \gamma^2 \delta \approx 4$.
This will ensure that an increase in $\phi$ by 1 will multiply the average model size in the search space by 4.
Here we use a multiplier of 4 (instead of 2) to spread out our search spaces so that we can run the same search space with multiple efficiency targets and sample our desired target model size from it.

As a result, our parameters for a given search space is the following:
\begin{align*}
    \text{depth: }& L \in \{d, \dots, 10d\} \\
    \text{base width: }& c^{\text{base}} \in \{16w, 24w, 48w, 96w, 96w, 192w\} \\
    \text{resolution: }& S = 224r \\
    \text{frame-rate: }& \tau = 5f.
\end{align*}
We round each of the above parameters to the nearest multiple of 8.
If $\phi = 0$, this forms the base search space for \ournet-A2.
Note that $c^{\text{expand}}$ is defined relative to $c^{\text{base}}$, so we do not need coefficients for it.

We found coefficients $\alpha, \beta, \gamma, \delta$ using a random search over these parameters.
More specifically, we select values in the range $[1.05, 1.40]$ at increments of 0.05 to represent possible values of the coefficients.
We ensure that the choice of coefficients is such that $\alpha \beta^2 \gamma^2 \delta \approx 4$, where the initial computation target for each frame is 300 MFLOPs.
For each combination, we scale the search space by the coefficients where $\phi = 1$, and randomly sample three architectures from each search space.
We train models for a selected search space for 10 epochs, averaging the results of the accuracy for that search space.
Then we select the coefficients that maximize the resulting accuracy.
Instead of selecting a single set of coefficients, we average the top 5 candidates to produce the final coefficients.
While the sample size of models is small and would be prone to noise, we find that the small averages work well in practice.

\subsubsection{Search Algorithm}

During search, we train a one-shot model using TuNAS~\cite{bender2020can} that overlaps all possible architectures into a hypernetwork.
At every step during optimization, we alternate between learning the network weights and learning a policy $\pi$ which we use to randomly sample a path through the hypernetwork to produce an initially random network architecture.
$\pi$ is learned using REINFORCE~\cite{williams1992simple}, optimized on the quality of sampled architectures, defined as the absolute reward consisting of the sampled network's accuracy and cost.
At each stage, the RL controller must choose a single categorical decision to select an architectural component.
The network architecture is a result of binding a value to each decision.
For example, the decision might choose between a spatial 1x3x3 convolution and a temporal 5x1x1 convolution.
We use FLOPs as the cost metric for architecture search, and use Kinetics 600 as the dataset to optimize for efficient video networks. During search, we obtain validation set accuracies on a held-out subset of the Kinetics 600 training set, training for a total of 90 epochs.

The addition of SE~\cite{hu2018squeeze} to our search space increases FLOPs by such a small amount ($<0.1$\%) that the search enables it for all layers.
SE plays a similar role as the feature gating in S3D-G~\cite{xie2017rethinking}, except with a nonlinear squeeze inside the projection operation.

\subsubsection{Stream Buffers for NAS}
We apply the stream buffers to \ournets as a separate step after NAS in the main text.
We can also leverage them for NAS to reduce memory usage during search.
Memory poses one of the biggest challenges for NAS, as models are forced to use a limited number of frames and small batch sizes to be able to keep the models in memory during optimization.
While this does not prevent us from performing search outright, it requires the use of accelerators with very high memory requirements, requiring a high cost of entry.
To circumvent this, we can use stream buffers with a small clip size to reduce memory. 
As a result, we can increase the total embedded frames and increase the batch size to provide better model accuracy estimation while running NAS.
Table~\ref{table:nas-stream} provides an example of an experiment where the use of a stream buffer can reduce memory requirements in this manner.
Using a stream buffer, we can reduce the input size from a single clip of 16 frames to 2 clips of 8 frames each, and double the batch size.
This results in a relatively modest increase in memory, compared to not using the buffer where we can run into out-of-memory (OOM) issues.

We note that the values of $b$ in each layer influences the memory consumption of the model.
This is dependent entirely on the temporal kernel width of the 3D convolution.
If $k = 5$, then we only need to cache the last 4 frames. 
$b$ could be larger, but but it will result in extra frames we will discard, so we set it to the smallest value to conserve memory.
Therefore, it is not necessary to specify it directly with NAS, as NAS is only concerned with the kernel sizes.
However, we can add an objective to NAS to minimize memory usage, which will apply pressure to reduce the temporal kernel widths and therefore will indirectly affect the value of $b$ in each layer. 
Reducing memory consumption even further by keeping kernel sizes can be explored in future work.


\begin{table}[tbp]
    \newcommand{\frameinput}[2]{#1$\times$#2$^2$}
    \footnotesize
    \begin{center}
    \begin{tabularx}{\columnwidth}{@{}Xrrrrrr@{}}
    \toprule
        \sc Config & \sc Full Input & \sc Batch Size & Memory & \sc Top-1 \\
    \midrule
        No Buffer & \frameinput{16}{172} & 8 & 5.8 GB & 55.9  \\
        Buffer (8 frames) & \frameinput{16}{172} & 16 & 6.6 GB & 58.5 \\
        No Buffer & \frameinput{16}{172} & 16 & 10.4 GB & OOM  \\
    \bottomrule
    \end{tabularx}
    \end{center}
    \caption{
        {\bf Effect of Stream Buffer on NAS (Kinetics 600)}.
        We measure the effect of NAS on \ournet-A0 when using a stream buffer vs. without on the same input.
        By embedding half the input at a time, we can double the batch size to improve the average NAS held-out accuracy without significantly increasing GPU memory per device.
    }
    \label{table:nas-stream}
\end{table}

\subsection{Architectures of \ournets} \label{appendix:architectures}






See Tables~\ref{table:a0-architecture-appendix}, \ref{table:a1-architecture-appendix}, \ref{table:a2-architecture-appendix}, \ref{table:a3-architecture-appendix}, \ref{table:a4-architecture-appendix}, and \ref{table:a5-architecture-appendix} for the architecture definitions of \ournet A0-A5 (we move the tables to the final pages of the Appendices to reduce clutter).
For \ournet-A6, we ensemble architectures A4 and A5 using the strategy described in the main text, i.e., we train both models independently and apply an arithmetic mean on the logits during inference.
All layers of all models have SE layers enabled, so we remove this search hyperparameter from all tables for brevity.

\subsubsection{More Details of the Architectures and Training}

We apply additional changes to our architectures and model training to improve performance even further.
To improve convergence speed in searching and training we use ReZero \cite{bachlechner2020rezero} by applying zero-initialized learnable scalar weights that are multiplied with features before the final sum in a residual block.
We also apply skip connections that are traditionally used in ResNets, adding a 1x1x1 convolution in the first layer of each block which may change the base channels or downsample the input.
However, we modify this to be similar to ResNet-D \cite{he2019bag} where we apply 1x3x3 spatial average pooling before the convolution to improve feature representations.

We apply Polyak averaging \cite{polyak1992acceleration} to the weights after every optimization step, using an Exponential Moving Average (EMA) with decay 0.99.
We adopt the Hard Swish activation function, which is a variant of SiLU/Swish \cite{elfwing2018sigmoid, ramachandran2017searching} proposed by MobileNetV3 \cite{howard2019searching} that is friendlier to quantization and CPU inference. We use the RMSProp optimizer with momentum 0.9 and a base learning rate of 1.8. 
We train for 240 epochs with a batch size of 1024 with synchronized batch normalization on all datasets and decay the learning rate using a cosine learning rate schedule~\cite{loshchilov2016sgdr} with a linear warm-up of 5 epochs.

We use a softmax cross-entropy loss with label smoothing 0.1 during training, except for Charades where we apply sigmoid cross-entropy to handle the multiple-class labels per video.
For Charades, we aggregate predictions across frames similar to AssembleNet~\cite{ryoo2019assemblenet}, where we apply a softmax across frames before applying temporal global average pooling to find multiple action classes that may occur in different frames.

Some works also expand the resolution for inference. For instance, X3D-M trains with a $224^2$ resolution while evaluating $256^2$ when using spatial crops.
We evaluate all of our models on the same resolution as training to make sure the FLOPs per frame during inference is unchanged from training.

\begin{figure}[t]
    \begin{center}
    \includegraphics[width=1.0\linewidth]{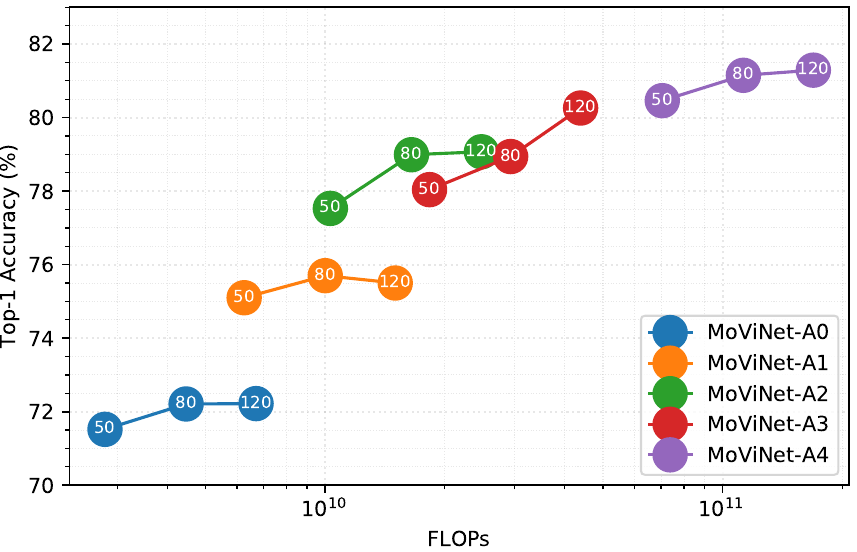}
    \end{center}
    \caption{
        {\bf Effect of Frame-Rate on Efficiency} on Kinetics 600.
        We train and evaluate each model with 50, 80, and 120 frames at 5fps, 8fps, and 12fps respectively.
    }
    \label{fig:frame-rate}
\end{figure}

Our choice of frame-rates can vary from model to model, providing different optimality depending on the architecture.
We plot the accuracy of training various \ournets on Kinetics 600 with different frame-rates in Figure~\ref{fig:frame-rate}.
Most models have good efficiency at 50 frames (5fps) or 80 frames (8fps) per video.
However, we can see \ournet-A4 benefits from a higher frame-rate of 12fps.
For Charades, we use 64 frames at 6fps for both training and inference.

\subsection{More Implementation Details and  Experiments}  \label{appendix:more-results}

\subsubsection{Implementing Causal Convolution by Padding}

\begin{figure}[t]
    \begin{center}
    \includegraphics[width=0.8\linewidth]{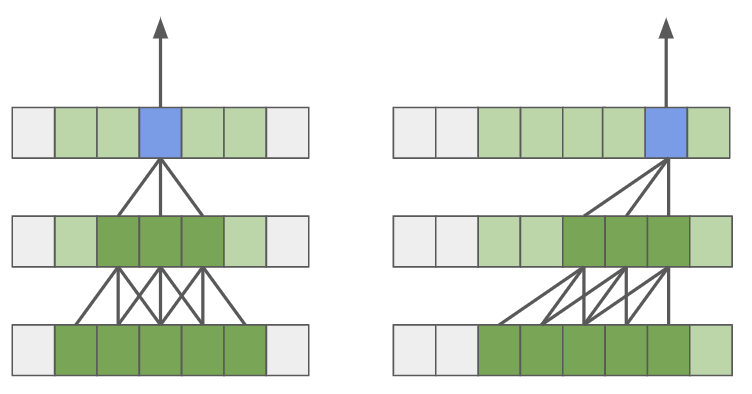}
    \end{center}
        \caption{
            {\bf Standard Convolution vs. Causal Convolution}. The figure illustrates the effective receptive field along a sequence of frames.
            The temporal kernel size is 3, with padding shown in white.
            Causal convolution can be performed efficiently by padding only on one side of the time axis thus to force the output causality.
        }
    \label{fig:causal-conv}
\end{figure}

To make a temporal convolution operation causal, we can apply a simple padding trick which shifts the receptive field forward such that the convolutional kernel is centered at the frame furthest into the future.
Figure~\ref{fig:causal-conv} illustrates this effect.
With a normal 3D convolution operation with kernel size $(k_t, k_h, k_w)$ and stride $s = 1$, the padding with respect to dimension $i$ is given as:
\begin{align}
    p_i^{\text{left}}, p_i^{\text{right}} =
    \begin{cases}
        (\frac{k_i-1}{2}, \frac{k_i-1}{2}) & \text{if $x$ is odd} \\
        (\frac{k_i-2}{2}, \frac{k_i}{2}) & \text{otherwise.}
    \end{cases}
\end{align}
where $p_i^{\text{left}}, p_i^{\text{right}}$ are the left and right padding amounts respectively.
For causal convolutions, we transform $p_t$ as:
\begin{align}
    p_t^{\text{left causal}}, p_t^{\text{right causal}} = (p_t^{\text{left}} + p_t^{\text{right}}, 0).
\end{align}
such that the effective temporal receptive field of a voxel at time position $t$ only spans $(0, t]$.

\subsubsection{Additional Details of Datasets}

We note that for all the Kinetics datasets are gradually shrinking over time due to videos being taken offline, making it difficult to compare against less recent works.
We report on the most recently available videos.
While putting our work at a disadvantage compared to previous work, we wish to make comparisons more fair for future work.
Nevertheless, we report the numbers as-is and report the reduction of examples in the datasets in Table~\ref{table:kinetics-num-examples}.

We report full results of all models in the following tables: Table~\ref{table:k400-comparison} for Kinetics 400, Table~\ref{table:k600-comparison-extra} for Kinetics 600 (with top-5 accuracy), Table~\ref{table:k700-comparison} for Kinetics 700, Table~\ref{table:mit-comparison} for Moments in Time, Table~\ref{table:charades-comparison} for Charades, Table~\ref{table:ssv2-comparison} for Something-Something V2, and Table~\ref{table:epic100-comparison} for Epic Kitchens 100.
For a table of results on Kinetics 600, see Table~\ref{table:k600-comparison-extra} and also the main text.

\begin{table}[tbp]
    \footnotesize
    \begin{center}
    \begin{tabularx}{\columnwidth}{@{}Xrrr@{}}
    \toprule
        \sc Dataset & \sc Train & \sc Valid & \sc Released \\
    \midrule
        Kinetics 400 & 215,435 (87.5\%) & 17,686 (88.4\%) & May 2017 \\
        Kinetics 600 & 364,305 (92.8\%) & 27,764 (92.5\%) & Aug 2018 \\
        Kinetics 700 & 524,595 (96.2\%) & 33,567 (95.9\%) & Jul 2019 \\
    \bottomrule
    \end{tabularx}
    \end{center}
    \caption{
        The number of examples available for each of the Kinetics dataset splits at the time of writing (Sept 20, 2020) along with the percentages compared to examples available on release.
        Each dataset loses about 4\% of its examples per year.
    }
    \label{table:kinetics-num-examples}
\end{table}

\begin{table}[tbp]
    \newcommand{\frameinput}[2]{#1$\times$#2$^2$}
    \begin{center}
    \footnotesize
    \begin{tabularx}{0.8\columnwidth}{@{}Xrrrr@{}}
    \toprule
        \sc Model & \sc Top-1 & \sc Top-5 & \sc gflops & \sc Param \\
    \midrule
        \bf \ournet-A0 & \bf 65.8 & \bf 87.4 & \bf 2.71 & \bf 3.1M \\
    \midrule
        \bf \ournet-A1 & \bf 72.7 & \bf 91.2 & \bf 6.02 & 4.6M \\
        X3D-XS \cite{feichtenhofer2020x3d} & 69.5 & - & 23.3 & \bf 3.8M \\
    \midrule
        \bf \ournet-A2 & \bf 75.0 & \bf 92.3 & \bf 10.3 & 4.8M \\
        X3D-S \cite{feichtenhofer2020x3d} & 73.5 & - & 76.1 & \bf 3.8M \\
    \midrule
        \bf \ournet-A3 & \bf 78.2 & \bf 93.8 & \bf 56.9 & 5.3M \\
        X3D-M \cite{feichtenhofer2020x3d} & 76.0 & 92.3 & \bf 186 & \bf 3.8M \\
    \midrule
        \bf \ournet-A4 & \bf 80.5 & \bf 94.5 & \bf 105 & \bf 5.9M \\
        X3D-L \cite{feichtenhofer2020x3d} & 77.5 & 92.9 & 744 & 6.1M \\
    \midrule
        \bf \ournet-A5 & \bf 80.9 & \bf 94.9 & \bf 281 & 15.7M \\
        X3D-XL \cite{feichtenhofer2020x3d} & 79.1 & 93.9 & 1452 & \bf 11.0M \\
    \midrule
        \bf \ournet-A6 & \bf 81.5 & \bf 95.3 & \bf 386 & 31.4M \\
        X3D-XXL~\cite{feichtenhofer2020x3d} & 80.4 & 94.6 & 5800 & \bf 20.3M \\
        TimeSformer-L~\cite{bertasius2021space} & 80.7 & 94.7 & 7140 & 120M \\
        ViViT-L/16x2~\cite{arnab2021vivit} & 81.3 & 94.7 & 3990 & 88.9M \\
    \bottomrule
    \end{tabularx}
    \end{center}
    \caption{
        {\bf Accuracy of \ournet on Kinetics 400}.
    }
    \label{table:k400-comparison}
\end{table}

\begin{table}[tbp]
    \begin{center}
    \resizebox{0.7\columnwidth}{!}{%
    \begin{tabular}{@{}lrrrr@{}}
    \toprule
        \sc Model & \sc Top-1 & Top-5 & \sc gflops & Param \\
    \midrule
        \bf \ournet-A0 & \bf 71.5 & \bf 90.4 & \bf 2.71 & \bf 3.1M \\
        \bf \ournet-A0-Stream & 70.3 & 90.1 & 2.73 & \bf 3.1M \\
    \midrule
        \bf \ournet-A1 & \bf 76.0 & 92.6 & \bf 6.02 & \bf 4.6M \\
        \bf \ournet-A1-Stream & 75.6 & \bf 92.8 & 6.06 & \bf 4.6M \\
    \midrule
        \bf \ournet-A2 & \bf 77.5 & \bf 93.4 & \bf 10.3 & \bf 4.8M \\
        \bf \ournet-A2-Stream & 76.5 & 93.3 & 10.4 & \bf 4.8M \\
    \midrule
        \bf \ournet-A3 & 80.8 & 94.5 & \bf 56.9 & \bf 5.3M \\
        \bf \ournet-A3 + AutoAugment & \bf 81.3 & \bf 95.3 & \bf 56.9 & \bf 5.3M \\
    \midrule
        \bf \ournet-A4 & 81.2 & 94.9 & \bf 105 & 4.9M \\
        \bf \ournet-A4  + AutoAugment & \bf 83.0 & \bf 96.0 & \bf 105 & 4.9M \\
        X3D-M \cite{feichtenhofer2020x3d} & 78.8 & 94.5 & 186 & \bf 3.8M \\
    \midrule
        \bf \ournet-A5 & 82.7 & 95.7 & \bf 281 & \bf 15.7M \\
        \bf \ournet-A5 + AutoAugment & \bf 84.3 & \bf 96.4 & \bf 281 & \bf 15.7M \\
    \midrule
        \bf \ournet-A6 & 83.5 & 96.2 & \bf 386 & 15.7M \\
        \bf \ournet-A6 + AutoAugment & \bf 84.8 & \bf 96.5 & \bf 386 & 15.7M \\
        X3D-XL \cite{feichtenhofer2020x3d} & 81.9 & 95.5 & 1452 & \bf 11.0M \\
        SlowFast-R50 \cite{feichtenhofer2019slowfast} & 78.8 & 94.0 & 1080 & 34.4M \\
        SlowFast-R101 \cite{feichtenhofer2019slowfast} & 81.8 & 95.1 & 7020 & 59.9M \\
    \bottomrule
    \end{tabular}
    }
    \end{center}
    \caption{
        {\bf Accuracy of \ournet on Kinetics 600} with additional top-5 data.
    }
    \label{table:k600-comparison-extra}
\end{table}

\begin{table}[tbp]
    \newcommand{\frameinput}[2]{#1$\times$#2$^2$}
    \begin{center}
    \footnotesize
    \begin{tabularx}{0.8\columnwidth}{@{}Xrrr@{}}
    \toprule
        \sc Model & \sc Top-1 & \sc gflops & \sc Param \\
    \midrule
        \bf \ournet-A0 & \bf 58.5 & \bf 2.71 & \bf 3.1M \\
    \midrule
        \bf \ournet-A1 & \bf 63.5 & \bf 6.02 & \bf 4.6M \\
    \midrule
        \bf \ournet-A2 & \bf 66.7 & \bf 10.3 & \bf 4.8M \\
    \midrule
        \bf \ournet-A3 & \bf 68.0 & \bf 56.9 & \bf 5.3M \\
    \midrule
        \bf \ournet-A4 & \bf 70.7 & \bf 105 & \bf 4.9M \\
    \midrule
        \bf \ournet-A5 & \bf 71.7 & \bf 281 & \bf 15.7M \\
    \midrule
        \bf \ournet-A6 & \bf 72.3 & \bf 386 & \bf 31.4M \\
        SlowFast-R101 \cite{feichtenhofer2019slowfast, activitynet2020} & 70.2 & 3200 & 30M \\
        SlowFast-R152 \cite{feichtenhofer2019slowfast, activitynet2020} & 71.6 & 9500 & 80M \\
    \midrule
        EfficientNet-L2 (pretrain) \cite{xie2020self, activitynet2020} & \bf 76.2 & \bf 15400 & \bf 480M \\
    \bottomrule
    \end{tabularx}
    \end{center}
    \caption{
        {\bf Accuracy of \ournet on Kinetics 700}.
    }
    \label{table:k700-comparison}
\end{table}

\begin{table}[tbp]
    \footnotesize
    \begin{center}
    \begin{tabularx}{\columnwidth}{@{}Xrrrr@{}}
    \toprule
        \sc Model & \sc Top-1 & \sc gflops & \sc Param \\
    \midrule
        \bf \ournet-A0 & \bf 27.5 & \bf 4.07 & \bf 3.1M \\
    \midrule
        \bf \ournet-A1 & \bf 32.0 & \bf 9.03 & \bf 4.6M \\
        TVN-1 \cite{piergiovanni2020tiny} & 23.1 & 13.0 & 11.1M \\
    \midrule
        \bf \ournet-A2 & \bf 34.3 & \bf 15.5 & \bf 4.8M \\
        \bf \ournet-A2-Stream & 33.6 & 15.6 & \bf 4.8M \\
        TVN-2 \cite{piergiovanni2020tiny} & 24.2 & 17.0 & 110M \\
    \midrule
        \bf \ournet-A3 & \bf 35.6 & \bf 35.6 & \bf 5.3M \\
        TVN-3 \cite{piergiovanni2020tiny} & 25.4 & 69.0 & 69.4M \\
    \midrule
        \bf \ournet-A4 & \bf 37.9 & \bf 98.4 & \bf 4.9M \\
        TVN-4 \cite{piergiovanni2020tiny} & 27.8 & \bf 106 & 44.2M \\
    \midrule
        \bf \ournet-A5 & \bf 39.1 & \bf 175 & \bf 15.7M \\
        SRTG-R3D-34 \cite{stergiou2020learn} & 28.5 & 220 & - \\
    \midrule
        \bf \ournet-A6 & \bf 40.2 & \bf 274 & \bf 31.4M \\
        ResNet3D-50 \cite{ryoo2019assemblenet} & 27.2 & - & - \\
        SRTG-R3D-50 \cite{stergiou2020learn} & 30.7 & 300 & - \\
        SRTG-R3D-101 \cite{stergiou2020learn} & 33.6 & 350 & - \\
        AssembleNet-50 (RGB+Flow) \cite{ryoo2019assemblenet} & 31.4 & 480 & 37.3M \\
        AssembleNet-101 (RGB+Flow) \cite{ryoo2019assemblenet} & 34.3 & 760 & 53.3M \\
        ViViT-L/16x2~\cite{arnab2021vivit} & 38.0 & 3410 & 100M \\
    \bottomrule
    \end{tabularx}
    \end{center}
    \caption{
        {\bf Accuracy of \ournet on Moments in Time}.
        All \ournets are evaluated on 75 frames at 25 fps.
    }
    \label{table:mit-comparison}
\end{table}

\begin{table}[tbp]
    \footnotesize
    \begin{center}
    \begin{tabularx}{\columnwidth}{@{}Xrrrr@{}}
    \toprule
        \sc Model & \sc mAP & \sc gflops & \sc Param \\
    \midrule
        \bf \ournet-A2 & \bf 32.5 & \bf 6.59 & \bf 4.8M \\
        TVN-1 \cite{piergiovanni2020tiny} & 32.2 & 13.0 & 11.1M \\
    \midrule
        TVN-2 \cite{piergiovanni2020tiny} & \bf 32.5 & \bf 17.0 & \bf 110M \\
    \midrule
        TVN-3 \cite{piergiovanni2020tiny} & \bf 33.5 & \bf 69.0 & \bf 69.4M \\
    \midrule
        \bf \ournet-A4 & \bf 48.5 & \bf 90.4 & \bf 4.9M \\
        TVN-4 \cite{piergiovanni2020tiny} & 35.4 & 106 & 44.2M \\
    \midrule
        \bf \ournet-A6 & \bf 63.2 & \bf 306 & \bf 31.4M \\
        AssembleNet-50 (RGB+Flow) \cite{ryoo2019assemblenet} & 53.0 & 700 & 37.3M \\
        AssembleNet-101 (RGB+Flow) \cite{ryoo2019assemblenet} & 58.6 & 1200 & 53.3M \\
        AssembleNet++ (RGB+Flow+Seg)\cite{ryoo2020assemblenetpp} & 59.8 & 1300 & - \\
        SlowFast 16x8 R101 \cite{feichtenhofer2019slowfast} & 45.2 & 7020 & 59.9M \\
    \bottomrule
    \end{tabularx}
    \end{center}
    \caption{
        {\bf Accuracy of \ournet on Charades}.
    }
    \label{table:charades-comparison}
\end{table}

\begin{table}[tbp]
    \begin{center}
    \resizebox{0.8\columnwidth}{!}{%
    \begin{tabular}{@{}lrrrr@{}}
    \toprule
        \sc Model & \sc Top-1 & Top-5 & \sc gflops & Param \\
    \midrule
        \bf \ournet-A0 & \bf 61.3 & 88.2 & \bf 2.71 & \bf 3.1M \\
        \bf \ournet-A0-Stream & 60.9 & \bf 88.3 & 2.73 & \bf 3.1M \\
        TRN~\cite{zhou2018temporal} & 48.8 & 77.6 & 33.0 & - \\
    \midrule
        \bf \ournet-A1 & \bf 62.7 & \bf 89.0 & \bf 6.02 & \bf 4.6M \\
        \bf \ournet-A1-Stream & 61.6 & 87.3 & 6.06 & \bf 4.6M \\
    \midrule
        \bf \ournet-A2 & \bf 63.5 & \bf 89.0 & \bf 10.3 & 4.8M \\
        \bf \ournet-A2-Stream & 63.1 & 89.0 & 10.4 & 4.8M \\
        TSM~\cite{lin2019tsm} & 63.4 & 88.5 & 390 & 24.3M \\
        VoV3D-M (16 frame)~\cite{lee2020diverse} & 63.2 & 88.2 & 34.2 & \bf 3.3M \\
    \midrule
        \bf \ournet-A3 & \bf 64.1 & \bf 88.8 & \bf 23.7 & \bf 5.3M \\
    \midrule
        VoV3D-M (32 frame)~\cite{lee2020diverse} & 65.2 & 89.4 & \bf 69.0 & \bf 3.3M \\
        VoV3D-L (32 frame)~\cite{lee2020diverse} & \bf 67.3 & \bf 90.5 & 125 & 5.8M \\
        ViViT-L/16x2~\cite{arnab2021vivit} & 65.4 & 89.8 & 3410 & 100M \\
    \bottomrule
    \end{tabular}
    }
    \end{center}
    \caption{
        {\bf Accuracy of \ournet on Something-Something V2}.
        All \ournets are evaluated on 50 frames at 12 fps.
        For shorter clips with fewer than 50 frames, we repeat the video sequence from the beginning.
    }
    \label{table:ssv2-comparison}
\end{table}

\begin{table}[tbp]
    \newcommand{\frameinput}[2]{#1$\times$#2$^2$}
    \begin{center}
    \footnotesize
    \begin{tabularx}{1\columnwidth}{@{}Xrrrrr@{}}
    \toprule
        \sc Model & \sc Action & \sc Verb & \sc Noun & \sc gflops & \sc Param \\
    \midrule
        \bf \ournet-A0 & \bf 36.8 & \bf 64.8 & \bf 47.4 & \bf 1.74 & \bf 3.1M \\
    \midrule
        \bf \ournet-A2 & \bf 41.2 & \bf 67.1 & \bf 52.3 & \bf 7.59 & \bf 4.8M \\
    \midrule
        \bf \ournet-A4 & \bf 44.4 & \bf 68.8 & \bf 56.2 & \bf 42.2 & \bf 4.9M \\
    \midrule
        \bf \ournet-A5 & \bf 44.5 & \bf 69.1 & \bf 55.1 & \bf 74.9 & \bf 15.7M \\
    \midrule
        \bf \ournet-A6 & \bf 47.7 & \bf 72.2 & \bf 57.3 & \bf 117 & \bf 31.4M \\
        ViViT-L/16x2~\cite{arnab2021vivit} & 44.0 & 66.4 & 56.8 & 3410 & 100M \\
        TSM~\cite{lin2019tsm} & 38.3 & 67.9 & 49.0 & - & - \\
        SlowFast~\cite{feichtenhofer2019slowfast} & 38.5 & 65.6 & 50.0 & - & - \\
        TSN~\cite{wang2016temporal} & 33.2 & 60.2 & 46.0 & - & - \\
    \bottomrule
    \end{tabularx}
    \end{center}
    \caption{
        {\bf Top-1 Accuracy of \ournet on Epic Kitchens 100} on Action, Verb, and Noun classes.
        All \ournets are evaluated on 32 frames at 12 fps.
    }
    \label{table:epic100-comparison}
\end{table}

\subsubsection{Single-Clip vs. Multi-Clip Evaluation}

We report all of our results on a \emph{single} view without multi-clip evaluation.
Additionally, we report the total number of frames used for evaluation and the frame rate (note that the evaluation frames can exceed the total number of frames in the reference video when subclips overlap).




As seen in Figure~1 and Table~2 (in the main text), switching from a multi-clip to single-clip X3D model on Kinetics 600 (where we cover the entire 10-second clip) results in much higher computational efficiency per video.
Existing work typically factors out FLOPs in terms of FLOPs per subclip, but it can hide the true cost of computation, since we can keep adding more clips to boost accuracy higher.

\begin{figure}[t]
    \begin{center}
    \includegraphics[width=1.0\linewidth]{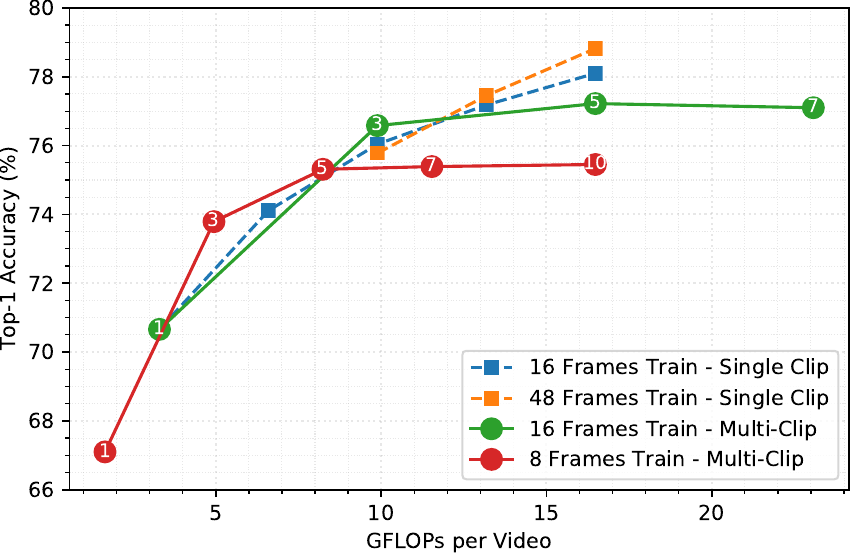}
    \end{center}
    \caption{
        {\bf Single vs. Multi-Clip Evaluation} on Kinetics 600.
        A comparison between the number of training frames and the number of eval frames and clips.
        The number of clips are shown inside each datapoint, where applicable.
        Other datapoints are evaluated on single clips.
        We use \ournet-A2 with frame stride 3 for all datapoints.
    }
    \label{fig:single-vs-multi-clip}
\end{figure}

We also evaluate the differences between training the same \ournet-A2 model on smaller clips vs. longer clips and evaluating the models with multi-clip vs. single-clip, as seen in Figure~\ref{fig:single-vs-multi-clip}.
For multi-clip evaluation, we can see that accuracy improves when the number of clips fill the whole duration of the video (this can be seen at 5 clips for 8 training frames and at 3 clips for 16 training frames), and only very slightly improves as we add more clips.
However, if we train \ournet-A2 on 16 frames and evaluate on 80 frames (so that we cover all 10 seconds of the video), this results in higher accuracy than the same number of frames using multi-clip eval.
Furthermore, we can boost this accuracy even higher if we use 48 frames to train our model.
Using stream buffers, we can reduce memory usage of training so that we can train using 48 frames while only using the memory of embedding 16 frames at a time.

\subsubsection{Streaming vs. Non-Streaming Evaluation}

\begin{figure}[t]
    \begin{center}
    \includegraphics[width=1.0\columnwidth]{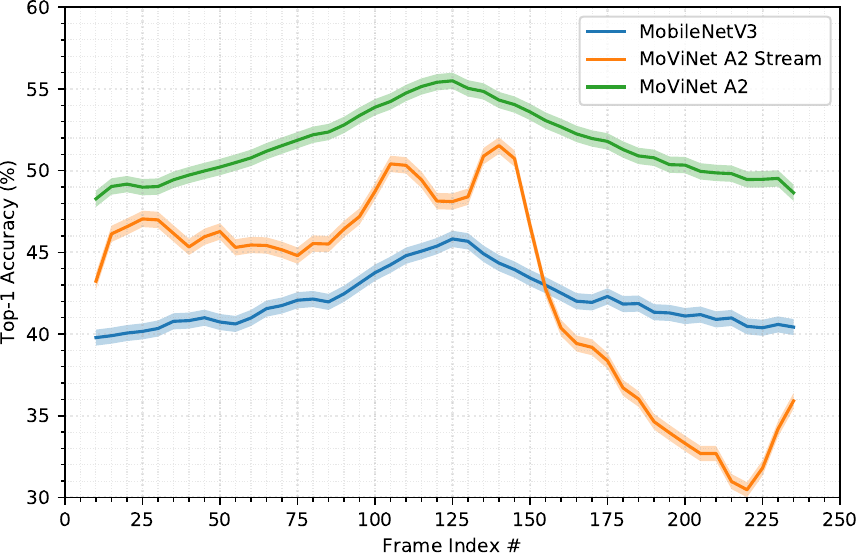}
    \end{center}
    \caption{
        {\bf Difference Between Streaming and Base MoViNets}.
        The plot displays the average accuracy across the Kinetics 600 dataset of an embedded model by applying the classification layers independently on each frame.
        Shading around each solid line indicates one standard deviation.
    }
    \label{fig:frame-stream-pool}
\end{figure}

One question we have wondered is if the distribution of features learned is different from streaming and non-streaming architectures.
In Figure~\ref{fig:frame-stream-pool}, we plot the average accuracy across Kinetics 600 of a model evaluated on a single frame by embedding an entire video, pooling across spatial dimensions, and applying the classification layers independently on each frame.

We first notice that the accuracy MobileNetV3 and MoViNet-A2 exhibit a Laplace distribution, on average peaking at the center frame of each video.
Since MobileNetV3 is evaluated on each frame independently, we can observe that the most salient part of the actions is on average in the video's midpoint.
This is a good indicator that the videos in Kinetics are trimmed very well to center around the most salient part of each action.
Likewise, MoViNet-A2, with balanced 3D convolutions, has the same characteristics as MobileNetV3, just with higher accuracy.

However, the dynamics of streaming MoViNet-A2 with causal convolutions is entirely different.
The distribution of accuracy fluctuates and varies more than non-streaming architectures.
By removing the ability for the network to see all frames as a whole with causal convolutions, the aggregation of features is not the same as when using balanced convolutions.
Despite this difference, overall, the accuracy difference across all videos is only about 1\%.
And by looking at top-5 accuracy in Table~\ref{table:k600-comparison-extra}, we can notice that streaming architectures nearly perform the same, despite the apparent information loss when transitioning to a model with a time-unidirectional receptive field.

\subsubsection{Long Video Sequences}

\begin{figure}[t]
    \begin{center}
    \includegraphics[width=1.0\linewidth]{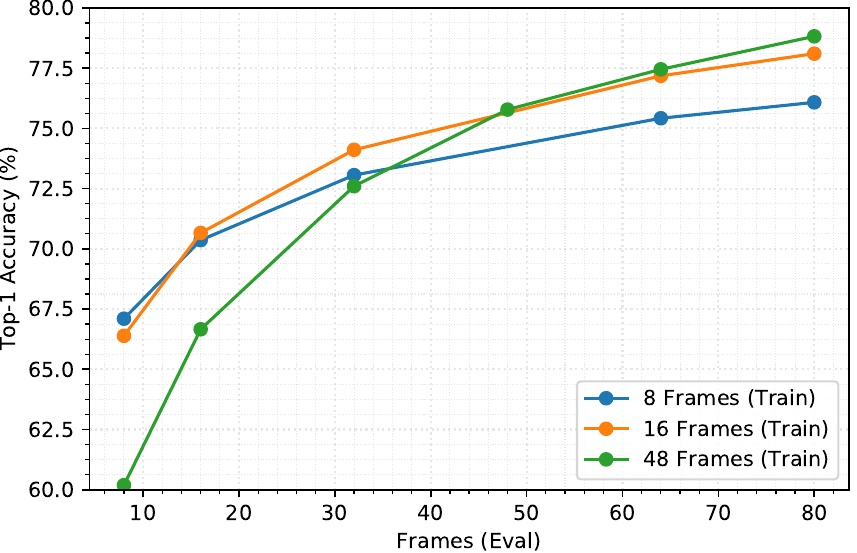}
    \end{center}
    \caption{
        {\bf Generalization to Longer Clips}. 
        A display of how duration of a clip during training affects the evaluation accuracy of different clip durations during evaluation. 
        We use \ournet-A2 with frame stride 3 for all datapoints.
    }
    \label{fig:frame-generalization}
\end{figure}

Figure~\ref{fig:frame-generalization} shows how training clip duration affects the accuracy of a model evaluated at different durations.
We can see that \ournet can generalize well beyond the original clip duration it was trained with, always improving in accuracy with more frames.
However, the model does notably worse if evaluated on clips with shorter durations than it was trained on.
Longer clip duration for training translates to better accuracy for evaluation on longer clips overall.
And with a stream buffer, we can train on even longer sequences to boost evaluation performance even higher.

However, we also see we can operate frame-by-frame with stream buffers, substantially saving memory, showing better memory efficiency than multi-clip approaches and requiring constant memory as the number of input frames increase (and therefore temporal receptive field).
Despite the accuracy reduction, we can see \ournet-Stream models perform very well on long video sequences and are still more efficient than X3D which requires splitting videos into smaller subclips.
We encourage future work using multi-clip evaluation to report results without overlapping subclips, which not only provides a much more representative accuracy measurement, tends to be more efficient as well.

\begin{table}[tbp]
    \footnotesize
    \begin{center}
    \begin{tabularx}{\columnwidth}{@{}Xrrr@{}}
    \toprule
        \sc Model & \sc Top-1 & \sc gflops & \sc Params \\
    \midrule
        \ournet-A2b-3D & 79.0 & 17.1 & 4.8M \\
        \ournet-A2b-(2+1)D & 79.4 & 16.8 & 5.0M \\
    \bottomrule
    \end{tabularx}
    \end{center}
    \caption{
        {\bf 3D vs. (2+1)D} on Kinetics 600.
    }
    \label{table:2p1d-vs-3d}
\end{table}

\subsubsection{Stream Buffers with Other Operations}

WaveNet \cite{oord2016wavenet} introduces causal convolution, where the receptive field on a stack of 1D convolutions is forced to only see activations up to the current time step, as opposed to balanced convolutions which expand their receptive fields in both directions.
We take inspiration from causal convolutions~\cite{cheng2019sparse,chang2018temporal,daiya2020stock} to design stream buffers.
However, WaveNet only proposes 1D convolutions for generative modeling, using them for their autoregressive property.
We generalize the idea of causal convolution to any local operation, and introduce stream buffers to be able to use causal operations for online inference, allowing frame-by-frame predictions.
In addition, Transformer-XL \cite{dai2019transformer} caches activations in a temporal buffer much like our work, for use in long-range sequence modeling.
However, the model is only causal across fixed sequences while our work can be causal across individual frames, and can even vary the number of frames in each clip (so long as frames are consecutive with no gaps or overlaps between clips).
We can apply the same principle to other operations as well to generalize causal operations.
Note that this approach is not inherently tied to any data type or modality.
Stream buffers can also be used to model many kinds of temporal data, e.g., audio, text.



\paragraph{(2+1)D CNNs.}
Additionally, support for efficient 3D convolutions on mobile devices is currently fragmented, while 2D convolutions are well supported.
We include the option to search for (2+1)D architectures, splitting up any 3D depthwise convolutions into a 2D spatial convolution followed by a 1D temporal convolution.
We show that trivially changing a 3D architecture to (2+1)D decreases FLOPs while also keeping similar accuracy, as seen in table~\ref{table:2p1d-vs-3d}.
Here we define \ournet-A2b as a searched model similar to \ournet-A2.

\newpage

\begin{table}[t]
    \newcommand{\baseframes}{50}
    
    \newcommand{\blockseq}[3]{\text{#1$\times$#2$^\text{2}$, #3}\\[-.1em]}
    \newcommand{\bigblock}[5]{
        block$_{#1}$ & 
        \multirow{#3}{*}{
            \(\left[
            \begin{array}{c}
                #4
            \end{array}
            \right]\)
        } &
        \multirow{#3}{*}{$\baseframes \times#2^\text{2}$} \\
        #5
    }
    \newcommand{\stemblock}[4]{
        data & stride #1, RGB & $\baseframes \times #3^\text{2}$ \\
        conv$_1$ & \multicolumn{1}{c}{$1\times3^\text{2}$, {#2}}
        & $\baseframes \times #4^\text{2}$ \\
    }
    \newcommand{\headblock}[3]{
        conv$_7$ & \multicolumn{1}{c}{$1\times1^\text{2}$, {#2}}
        & $50\times#1^\text{2}$ \\
        pool$_8$ & \multicolumn{1}{c}{$50\times#1^\text{2}$}
        & $1\times1^\text{2}$ \\
        dense$_9$ & \multicolumn{1}{c}{$1\times1^\text{2}$, {#3}}
        & $1\times1^\text{2}$ \\
        dense$_{10}$ & \multicolumn{1}{c}{$1\times1^\text{2}$, {600}}
        & $1\times1^\text{2}$ \\
    }

    \begin{center}
    \scriptsize
    \begin{tabularx}{0.8 \columnwidth}{@{}Xcl@{}}
        \toprule
        \sc Stage & \sc Operation & \sc Output size \\
        \midrule
        \stemblock{5}{8}{172}{86}
        \midrule
        \bigblock{2}{43}{2}{
            \blockseq{{1}}{{5}}{{8, 40}}
        }{
            &  & \\
        }
        \bigblock{3}{21}{3}{
            \blockseq{{5}}{{3}}{{32, 80}}
            \blockseq{{3}}{{3}}{{32, 80}}
            \blockseq{{3}}{{3}}{{32, 80}}
        }{
            &  & \\
            &  & \\
        }
        \bigblock{4}{10}{3}{
            \blockseq{{5}}{{3}}{{56, 184}}
            \blockseq{{3}}{{3}}{{56, 112}}
            \blockseq{{3}}{{3}}{{56, 184}}
        }{
            &  & \\
            &  & \\
        }
        \bigblock{5}{10}{4}{
            \blockseq{{5}}{{3}}{{56, 184}}
            \blockseq{{3}}{{3}}{{56, 184}}
            \blockseq{{3}}{{3}}{{56, 184}}
            \blockseq{{3}}{{3}}{{56, 184}}
        }{
            &  & \\
            &  & \\
            &  & \\
        }
        \bigblock{6}{5}{4}{
            \blockseq{{5}}{{3}}{{104, 344}}
            \blockseq{{1}}{{5}}{{104, 280}}
            \blockseq{{1}}{{5}}{{104, 280}}
            \blockseq{{1}}{{5}}{{104, 344}}
        }{
            &  & \\
            &  & \\
            &  & \\
        }
        \midrule
        \headblock{5}{480}{2048}
        \bottomrule
    \end{tabularx}
    \end{center}
    \caption{
        {\bf \ournet-A0 Architecture}.
    }
    \label{table:a0-architecture-appendix}
\end{table}

\begin{table}[t]
    \newcommand{\baseframes}{50}
    
    \newcommand{\blockseq}[3]{\text{#1$\times$#2$^\text{2}$, #3}\\[-.1em]}
    \newcommand{\bigblock}[5]{
        block$_{#1}$ & 
        \multirow{#3}{*}{
            \(\left[
            \begin{array}{c}
                #4
            \end{array}
            \right]\)
        } &
        \multirow{#3}{*}{$\baseframes \times#2^\text{2}$} \\
        #5
    }
    \newcommand{\stemblock}[4]{
        data & stride #1, RGB & $\baseframes \times #3^\text{2}$ \\
        conv$_1$ & \multicolumn{1}{c}{$1\times3^\text{2}$, {#2}}
        & $\baseframes \times #4^\text{2}$ \\
    }
    \newcommand{\headblock}[3]{
        conv$_7$ & \multicolumn{1}{c}{$1\times1^\text{2}$, {#2}}
        & $50\times#1^\text{2}$ \\
        pool$_8$ & \multicolumn{1}{c}{$50\times#1^\text{2}$}
        & $1\times1^\text{2}$ \\
        dense$_9$ & \multicolumn{1}{c}{$1\times1^\text{2}$, {#3}}
        & $1\times1^\text{2}$ \\
        dense$_{10}$ & \multicolumn{1}{c}{$1\times1^\text{2}$, {600}}
        & $1\times1^\text{2}$ \\
    }

    \begin{center}
    \scriptsize
    \begin{tabularx}{0.8 \columnwidth}{@{}Xcl@{}}
        \toprule
        \sc Stage & \sc Operation & \sc Output size \\
        \midrule
        \stemblock{5}{16}{172}{86}
        \midrule
        \bigblock{2}{43}{2}{
            \blockseq{{1}}{{5}}{{16, 40}}
            \blockseq{{3}}{{3}}{{16, 40}}
        }{
            &  & \\
        }
        \bigblock{3}{21}{4}{
            \blockseq{{3}}{{3}}{{40, 96}}
            \blockseq{{3}}{{3}}{{40, 120}}
            \blockseq{{3}}{{3}}{{40, 96}}
            \blockseq{{3}}{{3}}{{40, 96}}
        }{
            &  & \\
            &  & \\
            &  & \\
        }
        \bigblock{4}{10}{5}{
            \blockseq{{5}}{{3}}{{64, 216}}
            \blockseq{{3}}{{3}}{{64, 128}}
            \blockseq{{3}}{{3}}{{64, 216}}
            \blockseq{{3}}{{3}}{{64, 168}}
            \blockseq{{3}}{{3}}{{64, 216}}
        }{
            &  & \\
            &  & \\
            &  & \\
            &  & \\
        }
        \bigblock{5}{10}{6}{
            \blockseq{{5}}{{3}}{{64, 216}}
            \blockseq{{3}}{{3}}{{64, 216}}
            \blockseq{{3}}{{3}}{{64, 216}}
            \blockseq{{3}}{{3}}{{64, 128}}
            \blockseq{{1}}{{5}}{{64, 128}}
            \blockseq{{3}}{{3}}{{64, 216}}
        }{
            &  & \\
            &  & \\
            &  & \\
            &  & \\
            &  & \\
        }
        \bigblock{6}{5}{7}{
            \blockseq{{5}}{{3}}{{136, 456}}
            \blockseq{{1}}{{5}}{{136, 360}}
            \blockseq{{1}}{{5}}{{136, 360}}
            \blockseq{{1}}{{5}}{{136, 360}}
            \blockseq{{1}}{{5}}{{136, 456}}
            \blockseq{{3}}{{3}}{{136, 456}}
            \blockseq{{1}}{{3}}{{136, 544}}
        }{
            &  & \\
            &  & \\
            &  & \\
            &  & \\
            &  & \\
            &  & \\
        }
        \midrule
        \headblock{5}{600}{2048}
        \bottomrule
    \end{tabularx}
    \end{center}
    \caption{
        {\bf \ournet-A1 Architecture}.
    }
    \label{table:a1-architecture-appendix}
\end{table}

\begin{table}[t]
    \newcommand{\baseframes}{50}
    
    \newcommand{\blockseq}[3]{\text{#1$\times$#2$^\text{2}$, #3}\\[-.1em]}
    \newcommand{\bigblock}[5]{
        block$_{#1}$ & 
        \multirow{#3}{*}{
            \(\left[
            \begin{array}{c}
                #4
            \end{array}
            \right]\)
        } &
        \multirow{#3}{*}{$\baseframes \times#2^\text{2}$} \\
        #5
    }
    \newcommand{\stemblock}[4]{
        data & stride #1, RGB & $\baseframes \times #3^\text{2}$ \\
        conv$_1$ & \multicolumn{1}{c}{$1\times3^\text{2}$, {#2}}
        & $\baseframes \times #4^\text{2}$ \\
    }
    \newcommand{\headblock}[3]{
        conv$_7$ & \multicolumn{1}{c}{$1\times1^\text{2}$, {#2}}
        & $50\times#1^\text{2}$ \\
        pool$_8$ & \multicolumn{1}{c}{$50\times#1^\text{2}$}
        & $1\times1^\text{2}$ \\
        dense$_9$ & \multicolumn{1}{c}{$1\times1^\text{2}$, {#3}}
        & $1\times1^\text{2}$ \\
        dense$_{10}$ & \multicolumn{1}{c}{$1\times1^\text{2}$, {600}}
        & $1\times1^\text{2}$ \\
    }

    \begin{center}
    \scriptsize
    \begin{tabularx}{0.8 \columnwidth}{@{}Xcl@{}}
        \toprule
        \sc Stage & \sc Operation & \sc Output size \\
        \midrule
        \stemblock{5}{16}{224}{112}
        \midrule
        \bigblock{2}{56}{3}{
            \blockseq{{1}}{{5}}{{16, 40}}
            \blockseq{{3}}{{3}}{{16, 40}}
            \blockseq{{3}}{{3}}{{16, 64}}
        }{
            &  & \\
            &  & \\
        }
        \bigblock{3}{28}{5}{
            \blockseq{{3}}{{3}}{{40, 96}}
            \blockseq{{3}}{{3}}{{40, 120}}
            \blockseq{{3}}{{3}}{{40, 96}}
            \blockseq{{3}}{{3}}{{40, 96}}
            \blockseq{{3}}{{3}}{{40, 120}}
        }{
            &  & \\
            &  & \\
            &  & \\
            &  & \\
        }
        \bigblock{4}{14}{5}{
            \blockseq{{5}}{{3}}{{72, 240}}
            \blockseq{{3}}{{3}}{{72, 160}}
            \blockseq{{3}}{{3}}{{72, 240}}
            \blockseq{{3}}{{3}}{{72, 192}}
            \blockseq{{3}}{{3}}{{72, 240}}
        }{
            &  & \\
            &  & \\
            &  & \\
            &  & \\
        }
        \bigblock{5}{14}{6}{
            \blockseq{{5}}{{3}}{{72, 240}}
            \blockseq{{3}}{{3}}{{72, 240}}
            \blockseq{{3}}{{3}}{{72, 240}}
            \blockseq{{3}}{{3}}{{72, 240}}
            \blockseq{{1}}{{5}}{{72, 144}}
            \blockseq{{3}}{{3}}{{72, 240}}
        }{
            &  & \\
            &  & \\
            &  & \\
            &  & \\
            &  & \\
        }
        \bigblock{6}{7}{7}{
            \blockseq{{5}}{{3}}{{144, 480}}
            \blockseq{{1}}{{5}}{{144, 384}}
            \blockseq{{1}}{{5}}{{144, 384}}
            \blockseq{{1}}{{5}}{{144, 480}}
            \blockseq{{1}}{{5}}{{144, 480}}
            \blockseq{{3}}{{3}}{{144, 480}}
            \blockseq{{1}}{{3}}{{144, 576}}
        }{
            &  & \\
            &  & \\
            &  & \\
            &  & \\
            &  & \\
            &  & \\
        }
        \midrule
        \headblock{7}{640}{2048}
        \bottomrule
    \end{tabularx}
    \end{center}
    \caption{
        {\bf \ournet-A2 Architecture}.
    }
    \label{table:a2-architecture-appendix}
\end{table}

\begin{table}[t]
    \newcommand{\baseframes}{120}
    
    \newcommand{\blockseq}[3]{\text{#1$\times$#2$^\text{2}$, #3}\\[-.1em]}
    \newcommand{\bigblock}[5]{
        block$_{#1}$ & 
        \multirow{#3}{*}{
            \(\left[
            \begin{array}{c}
                #4
            \end{array}
            \right]\)
        } &
        \multirow{#3}{*}{$\baseframes \times#2^\text{2}$} \\
        #5
    }
    \newcommand{\stemblock}[4]{
        data & stride #1, RGB & $\baseframes \times #3^\text{2}$ \\
        conv$_1$ & \multicolumn{1}{c}{$1\times3^\text{2}$, {#2}}
        & $\baseframes \times #4^\text{2}$ \\
    }
    \newcommand{\headblock}[3]{
        conv$_7$ & \multicolumn{1}{c}{$1\times1^\text{2}$, {#2}}
        & $\baseframes\times#1^\text{2}$ \\
        pool$_8$ & \multicolumn{1}{c}{$\baseframes\times#1^\text{2}$}
        & $1\times1^\text{2}$ \\
        dense$_9$ & \multicolumn{1}{c}{$1\times1^\text{2}$, {#3}}
        & $1\times1^\text{2}$ \\
        dense$_{10}$ & \multicolumn{1}{c}{$1\times1^\text{2}$, {600}}
        & $1\times1^\text{2}$ \\
    }

    \begin{center}
    \scriptsize
    \begin{tabularx}{0.8 \columnwidth}{@{}Xcl@{}}
        \toprule
        \sc Stage & \sc Operation & \sc Output size \\
        \midrule
        \stemblock{5}{16}{256}{128}
        \midrule
        \bigblock{2}{64}{4}{
            \blockseq{{1}}{{5}}{{16, 40}}
            \blockseq{{3}}{{3}}{{16, 40}}
            \blockseq{{3}}{{3}}{{16, 64}}
            \blockseq{{3}}{{3}}{{16, 40}}
        }{
            &  & \\
            &  & \\
            &  & \\
        }
        \bigblock{3}{32}{6}{
            \blockseq{{3}}{{3}}{{48, 112}}
            \blockseq{{3}}{{3}}{{48, 144}}
            \blockseq{{3}}{{3}}{{48, 112}}
            \blockseq{{1}}{{5}}{{48, 112}}
            \blockseq{{3}}{{3}}{{48, 144}}
            \blockseq{{3}}{{3}}{{48, 144}}
        }{
            &  & \\
            &  & \\
            &  & \\
            &  & \\
            &  & \\
        }
        \bigblock{4}{16}{5}{
            \blockseq{{5}}{{3}}{{80, 240}}
            \blockseq{{3}}{{3}}{{80, 152}}
            \blockseq{{3}}{{3}}{{80, 240}}
            \blockseq{{3}}{{3}}{{80, 192}}
            \blockseq{{3}}{{3}}{{80, 240}}
        }{
            &  & \\
            &  & \\
            &  & \\
            &  & \\
        }
        \bigblock{5}{16}{8}{
            \blockseq{{5}}{{3}}{{88, 264}}
            \blockseq{{3}}{{3}}{{88, 264}}
            \blockseq{{3}}{{3}}{{88, 264}}
            \blockseq{{3}}{{3}}{{88, 264}}
            \blockseq{{1}}{{5}}{{88, 160}}
            \blockseq{{3}}{{3}}{{88, 264}}
            \blockseq{{3}}{{3}}{{88, 264}}
            \blockseq{{3}}{{3}}{{88, 264}}
        }{
            &  & \\
            &  & \\
            &  & \\
            &  & \\
            &  & \\
            &  & \\
            &  & \\
        }
        \bigblock{6}{8}{10}{
            \blockseq{{5}}{{3}}{{168, 560}}
            \blockseq{{1}}{{5}}{{168, 448}}
            \blockseq{{1}}{{5}}{{168, 448}}
            \blockseq{{1}}{{5}}{{168, 560}}
            \blockseq{{1}}{{5}}{{168, 560}}
            \blockseq{{3}}{{3}}{{168, 560}}
            \blockseq{{1}}{{5}}{{168, 448}}
            \blockseq{{1}}{{5}}{{168, 448}}
            \blockseq{{3}}{{3}}{{168, 560}}
            \blockseq{{1}}{{3}}{{168, 672}}
        }{
            &  & \\
            &  & \\
            &  & \\
            &  & \\
            &  & \\
            &  & \\
            &  & \\
            &  & \\
            &  & \\
        }
        \midrule
        \headblock{8}{744}{2048}
        \bottomrule
    \end{tabularx}
    \end{center}
    \caption{
        {\bf \ournet-A3 Architecture}.
    }
    \label{table:a3-architecture-appendix}
\end{table}

\begin{table}[t]
    \newcommand{\baseframes}{80}
    
    \newcommand{\blockseq}[3]{\text{#1$\times$#2$^\text{2}$, #3}\\[-.1em]}
    \newcommand{\bigblock}[5]{
        block$_{#1}$ & 
        \multirow{#3}{*}{
            \(\left[
            \begin{array}{c}
                #4
            \end{array}
            \right]\)
        } &
        \multirow{#3}{*}{$\baseframes \times#2^\text{2}$} \\
        #5
    }
    \newcommand{\stemblock}[4]{
        data & stride #1, RGB & $\baseframes \times #3^\text{2}$ \\
        conv$_1$ & \multicolumn{1}{c}{$1\times3^\text{2}$, {#2}}
        & $\baseframes \times #4^\text{2}$ \\
    }
    \newcommand{\headblock}[3]{
        conv$_7$ & \multicolumn{1}{c}{$1\times1^\text{2}$, {#2}}
        & $\baseframes\times#1^\text{2}$ \\
        pool$_8$ & \multicolumn{1}{c}{$\baseframes\times#1^\text{2}$}
        & $1\times1^\text{2}$ \\
        dense$_9$ & \multicolumn{1}{c}{$1\times1^\text{2}$, {#3}}
        & $1\times1^\text{2}$ \\
        dense$_{10}$ & \multicolumn{1}{c}{$1\times1^\text{2}$, {600}}
        & $1\times1^\text{2}$ \\
    }

    \begin{center}
    \scriptsize
    \begin{tabularx}{0.8 \columnwidth}{@{}Xcl@{}}
        \toprule
        \sc Stage & \sc Operation & \sc Output size \\
        \midrule
        \stemblock{5}{24}{290}{145}
        \midrule
        \bigblock{2}{72}{6}{
            \blockseq{{1}}{{5}}{{24, 64}}
            \blockseq{{3}}{{3}}{{24, 64}}
            \blockseq{{3}}{{3}}{{24, 96}}
            \blockseq{{3}}{{3}}{{24, 64}}
            \blockseq{{3}}{{3}}{{24, 96}}
            \blockseq{{3}}{{3}}{{24, 64}}
        }{
            &  & \\
            &  & \\
            &  & \\
            &  & \\
            &  & \\
        }
        \bigblock{3}{36}{9}{
            \blockseq{{5}}{{3}}{{56, 168}}
            \blockseq{{3}}{{3}}{{56, 168}}
            \blockseq{{3}}{{3}}{{56, 136}}
            \blockseq{{3}}{{3}}{{56, 136}}
            \blockseq{{3}}{{3}}{{56, 168}}
            \blockseq{{3}}{{3}}{{56, 168}}
            \blockseq{{3}}{{3}}{{56, 168}}
            \blockseq{{1}}{{5}}{{56, 136}}
            \blockseq{{3}}{{3}}{{56, 136}}
        }{
            &  & \\
            &  & \\
            &  & \\
            &  & \\
            &  & \\
            &  & \\
            &  & \\
            &  & \\
        }
        \bigblock{4}{18}{9}{
            \blockseq{{5}}{{3}}{{96, 320}}
            \blockseq{{3}}{{3}}{{96, 160}}
            \blockseq{{3}}{{3}}{{96, 320}}
            \blockseq{{3}}{{3}}{{96, 192}}
            \blockseq{{3}}{{3}}{{96, 320}}
            \blockseq{{3}}{{3}}{{96, 152}}
            \blockseq{{3}}{{3}}{{96, 320}}
            \blockseq{{3}}{{3}}{{96, 256}}
            \blockseq{{3}}{{3}}{{96, 320}}
        }{
            &  & \\
            &  & \\
            &  & \\
            &  & \\
            &  & \\
            &  & \\
            &  & \\
            &  & \\
        }
        \bigblock{5}{18}{10}{
            \blockseq{{5}}{{3}}{{96, 320}}
            \blockseq{{3}}{{3}}{{96, 320}}
            \blockseq{{3}}{{3}}{{96, 320}}
            \blockseq{{3}}{{3}}{{96, 320}}
            \blockseq{{1}}{{5}}{{96, 192}}
            \blockseq{{3}}{{3}}{{96, 320}}
            \blockseq{{3}}{{3}}{{96, 320}}
            \blockseq{{3}}{{3}}{{96, 192}}
            \blockseq{{3}}{{3}}{{96, 320}}
            \blockseq{{3}}{{3}}{{96, 320}}
        }{
            &  & \\
            &  & \\
            &  & \\
            &  & \\
            &  & \\
            &  & \\
            &  & \\
            &  & \\
            &  & \\
        }
        \bigblock{6}{9}{13}{
            \blockseq{{5}}{{3}}{{192, 640}}
            \blockseq{{1}}{{5}}{{192, 512}}
            \blockseq{{1}}{{5}}{{192, 512}}
            \blockseq{{1}}{{5}}{{192, 640}}
            \blockseq{{1}}{{5}}{{192, 640}}
            \blockseq{{3}}{{3}}{{192, 640}}
            \blockseq{{1}}{{5}}{{192, 512}}
            \blockseq{{1}}{{5}}{{192, 512}}
            \blockseq{{1}}{{5}}{{192, 640}}
            \blockseq{{1}}{{5}}{{192, 768}}
            \blockseq{{1}}{{5}}{{192, 640}}
            \blockseq{{3}}{{3}}{{192, 640}}
            \blockseq{{3}}{{3}}{{192, 768}}
        }{
            &  & \\
            &  & \\
            &  & \\
            &  & \\
            &  & \\
            &  & \\
            &  & \\
            &  & \\
            &  & \\
            &  & \\
            &  & \\
            &  & \\
        }
        \midrule
        \headblock{9}{856}{2048}
        \bottomrule
    \end{tabularx}
    \end{center}
    \caption{
        {\bf \ournet-A4 Architecture}.
    }
    \label{table:a4-architecture-appendix}
\end{table}

\begin{table}[t]
    \newcommand{\baseframes}{120}
    
    \newcommand{\blockseq}[3]{\text{#1$\times$#2$^\text{2}$, #3}\\[-.1em]}
    \newcommand{\bigblock}[5]{
        block$_{#1}$ & 
        \multirow{#3}{*}{
            \(\left[
            \begin{array}{c}
                #4
            \end{array}
            \right]\)
        } &
        \multirow{#3}{*}{$\baseframes \times#2^\text{2}$} \\
        #5
    }
    \newcommand{\stemblock}[4]{
        data & stride #1, RGB & $\baseframes \times #3^\text{2}$ \\
        conv$_1$ & \multicolumn{1}{c}{$1\times3^\text{2}$, {#2}}
        & $\baseframes \times #4^\text{2}$ \\
    }
    \newcommand{\headblock}[3]{
        conv$_7$ & \multicolumn{1}{c}{$1\times1^\text{2}$, {#2}}
        & $\baseframes\times#1^\text{2}$ \\
        pool$_8$ & \multicolumn{1}{c}{$\baseframes\times#1^\text{2}$}
        & $1\times1^\text{2}$ \\
        dense$_9$ & \multicolumn{1}{c}{$1\times1^\text{2}$, {#3}}
        & $1\times1^\text{2}$ \\
        dense$_{10}$ & \multicolumn{1}{c}{$1\times1^\text{2}$, {600}}
        & $1\times1^\text{2}$ \\
    }

    \begin{center}
    \scriptsize
    \begin{tabularx}{0.8 \columnwidth}{@{}Xcl@{}}
        \toprule
        \sc Stage & \sc Operation & \sc Output size \\
        \midrule
        \stemblock{5}{24}{320}{160}
        \midrule
        \bigblock{2}{80}{6}{
            \blockseq{{1}}{{5}}{{24, 64}}
            \blockseq{{1}}{{5}}{{24, 64}}
            \blockseq{{3}}{{3}}{{24, 96}}
            \blockseq{{3}}{{3}}{{24, 64}}
            \blockseq{{3}}{{3}}{{24, 96}}
            \blockseq{{3}}{{3}}{{24, 64}}
        }{
            &  & \\
            &  & \\
            &  & \\
            &  & \\
            &  & \\
            &  & \\
        }
        \bigblock{3}{40}{11}{
            \blockseq{{5}}{{3}}{{64, 192}}
            \blockseq{{3}}{{3}}{{64, 152}}
            \blockseq{{3}}{{3}}{{64, 152}}
            \blockseq{{3}}{{3}}{{64, 152}}
            \blockseq{{3}}{{3}}{{64, 192}}
            \blockseq{{3}}{{3}}{{64, 192}}
            \blockseq{{3}}{{3}}{{64, 192}}
            \blockseq{{3}}{{3}}{{64, 152}}
            \blockseq{{3}}{{3}}{{64, 152}}
            \blockseq{{3}}{{3}}{{64, 192}}
            \blockseq{{3}}{{3}}{{64, 192}}
        }{
            &  & \\
            &  & \\
            &  & \\
            &  & \\
            &  & \\
            &  & \\
            &  & \\
            &  & \\
            &  & \\
            &  & \\
            &  & \\
        }
        \bigblock{4}{20}{13}{
            \blockseq{{5}}{{3}}{{112, 376}}
            \blockseq{{3}}{{3}}{{112, 224}}
            \blockseq{{3}}{{3}}{{112, 376}}
            \blockseq{{3}}{{3}}{{112, 376}}
            \blockseq{{3}}{{3}}{{112, 296}}
            \blockseq{{3}}{{3}}{{112, 376}}
            \blockseq{{3}}{{3}}{{112, 224}}
            \blockseq{{3}}{{3}}{{112, 376}}
            \blockseq{{3}}{{3}}{{112, 376}}
            \blockseq{{3}}{{3}}{{112, 296}}
            \blockseq{{3}}{{3}}{{112, 376}}
            \blockseq{{3}}{{3}}{{112, 376}}
            \blockseq{{3}}{{3}}{{112, 376}}
        }{
            &  & \\
            &  & \\
            &  & \\
            &  & \\
            &  & \\
            &  & \\
            &  & \\
            &  & \\
            &  & \\
            &  & \\
            &  & \\
            &  & \\
            &  & \\
        }
        \bigblock{5}{20}{11}{
            \blockseq{{5}}{{3}}{{120, 376}}
            \blockseq{{3}}{{3}}{{120, 376}}
            \blockseq{{3}}{{3}}{{120, 376}}
            \blockseq{{3}}{{3}}{{120, 376}}
            \blockseq{{1}}{{5}}{{120, 224}}
            \blockseq{{3}}{{3}}{{120, 376}}
            \blockseq{{3}}{{3}}{{120, 376}}
            \blockseq{{3}}{{3}}{{120, 224}}
            \blockseq{{3}}{{3}}{{120, 376}}
            \blockseq{{3}}{{3}}{{120, 376}}
            \blockseq{{3}}{{3}}{{120, 376}}
        }{
            &  & \\
            &  & \\
            &  & \\
            &  & \\
            &  & \\
            &  & \\
            &  & \\
            &  & \\
            &  & \\
            &  & \\
            &  & \\
        }
        \bigblock{6}{10}{18}{
            \blockseq{{5}}{{3}}{{224, 744}}
            \blockseq{{3}}{{3}}{{224, 744}}
            \blockseq{{1}}{{5}}{{224, 600}}
            \blockseq{{1}}{{5}}{{224, 600}}
            \blockseq{{1}}{{5}}{{224, 744}}
            \blockseq{{1}}{{5}}{{224, 744}}
            \blockseq{{3}}{{3}}{{224, 744}}
            \blockseq{{1}}{{5}}{{224, 896}}
            \blockseq{{1}}{{5}}{{224, 600}}
            \blockseq{{1}}{{5}}{{224, 600}}
            \blockseq{{1}}{{5}}{{224, 896}}
            \blockseq{{1}}{{5}}{{224, 744}}
            \blockseq{{3}}{{3}}{{224, 744}}
            \blockseq{{1}}{{5}}{{224, 896}}
            \blockseq{{1}}{{5}}{{224, 600}}
            \blockseq{{1}}{{5}}{{224, 600}}
            \blockseq{{1}}{{5}}{{224, 744}}
            \blockseq{{3}}{{3}}{{224, 744}}
        }{
            &  & \\
            &  & \\
            &  & \\
            &  & \\
            &  & \\
            &  & \\
            &  & \\
            &  & \\
            &  & \\
            &  & \\
            &  & \\
            &  & \\
            &  & \\
            &  & \\
            &  & \\
            &  & \\
            &  & \\
            &  & \\
        }
        \midrule
        \headblock{10}{992}{2048}
        \bottomrule
    \end{tabularx}
    \end{center}
    \caption{
        {\bf \ournet-A5 Architecture}.
    }
    \label{table:a5-architecture-appendix}
\end{table}

\end{document}